\documentclass{bmvc2k}

\usepackage{caption}
\usepackage{subcaption}
\usepackage{amsmath}
\usepackage{amssymb}
\usepackage{amsfonts} 
\usepackage{relsize}
\usepackage{multirow}
\usepackage{color,soul}

\title{Hierarchical Residual Learning Based Vector Quantized Image Modeling with VQVAE}
\title{Hierarchical Residual Learning Based Vector Quantized Variational Autoencoder for Image Reconstruction and Generation}

\addauthor{Mohammad Adiban}{mohammad.adiban@ntnu.no}{1,2}
\addauthor{Kalin Stefanov}{kalin.stefanov@monash.edu}{2}
\addauthor{Sabato Marco Siniscalchi}{marco.siniscalchi@ntnu.no}{1}
\addauthor{Giampiero Salvi}{giampiero.salvi@ntnu.no}{1,3}

\addinstitution{Norwegian University of Science and Technology\\Trondheim, Norway}
\addinstitution{Monash University\\Melbourne, Australia}
\addinstitution{KTH Royal Institute of Technology\\Stockholm, Sweden}

\runninghead{Adiban, Stefanov, Siniscalchi, Salvi}{HR-VQVAE}

\begin{document}

\maketitle

\begin{abstract}
We propose a multi-layer variational autoencoder method, we call HR-VQVAE, that learns hierarchical discrete representations of the data. By utilizing a novel objective function, each layer in HR-VQVAE learns a discrete representation of the residual from previous layers through a vector quantized encoder. Furthermore, the representations at each layer are hierarchically linked to those at previous layers. We evaluate our method on the tasks of image reconstruction and generation. Experimental results demonstrate that the discrete representations learned by HR-VQVAE enable the decoder to reconstruct high-quality images with less distortion than the baseline methods, namely VQVAE and VQVAE-2. HR-VQVAE can also generate high-quality and diverse images that outperform state-of-the-art generative models, providing further verification of the efficiency of the learned representations. The hierarchical nature of HR-VQVAE i)~reduces the decoding search time, making the method particularly suitable for high-load tasks and ii)~allows to increase the codebook size without incurring the codebook collapse problem.
\end{abstract}

%%% Introduction
\section{Introduction}
\label{sec:intro}
Deep generative modeling has shown impressive results for the application of unsupervised learning in many domains, e.g., image super-resolution~\cite{ledig2017photo}, image generation~\cite{theis2016note}, and future video frame prediction~\cite{liang2017dual}. %, and text-to-speech and music generation~\cite{oord2016wavenet}.
Variational autoencoders (VAEs) \cite{zhao2017towards}, which are the focus of this work, compute continuous-valued representations by compressing information into a dense, distributed embedding~\cite{zhao2017towards}. However, studies on human cognition emphasize the importance of discretization in representation learning. Discrete symbolic representations contribute to reasoning, understanding, generalization, and efficient learning~\cite{cartuyvels2021discrete}.
Discrete representations can also significantly reduce the computational complexity and improve interpretability by illustrating which terms contributed to the solution~\cite{mordatch2018emergence}. %These discrete representations can also facilitate the encoding of inductive biases in the learning process, such as images consisting of a small number of objects~\cite{eslami2016attend}.

\begin{figure}
    \centering
    \begin{subfigure}[b]{0.24\textwidth}
        \centering
        \includegraphics[width=\textwidth]{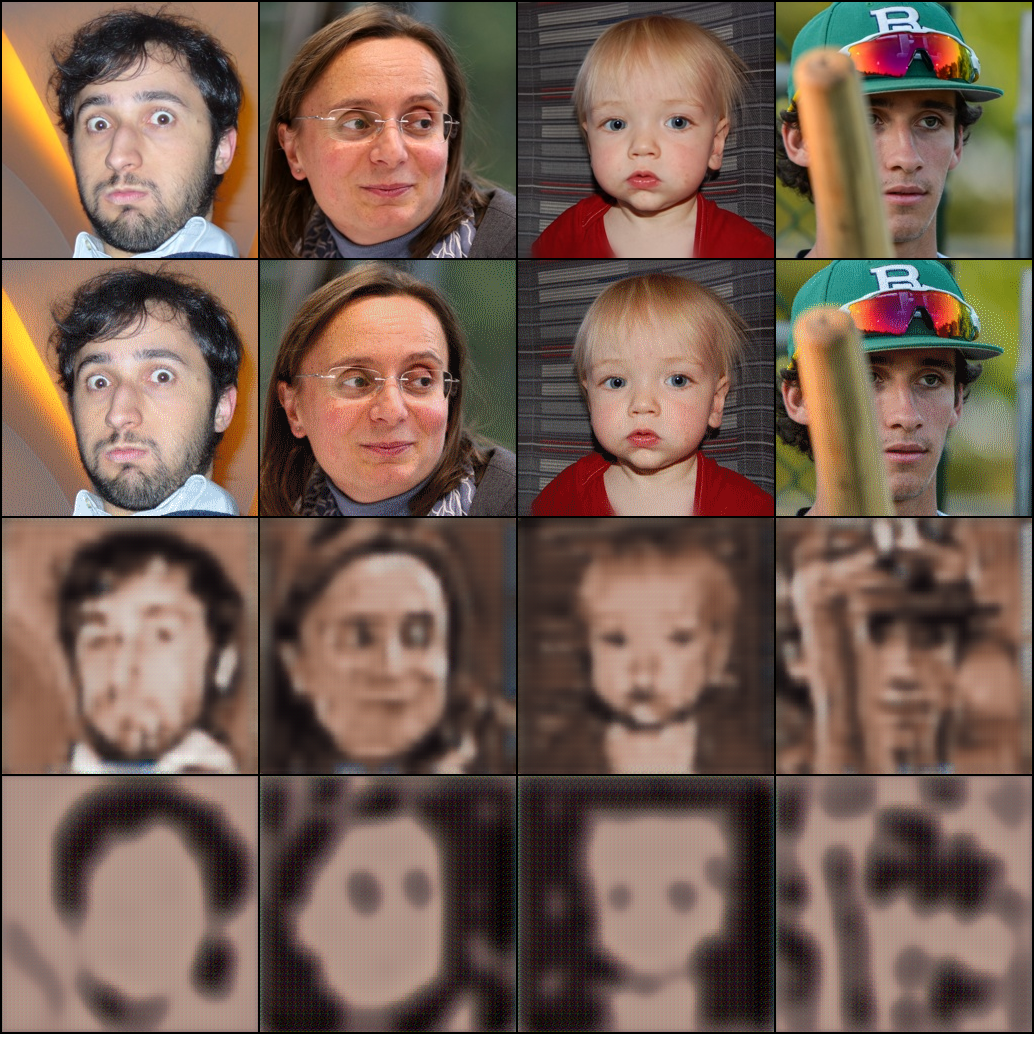}
        \caption{{\small FFHQ}}
        \label{fig:ffhq_recon}
    \end{subfigure}
    \hfill
    \begin{subfigure}[b]{0.24\textwidth}  
        \centering 
        \includegraphics[width=\textwidth]{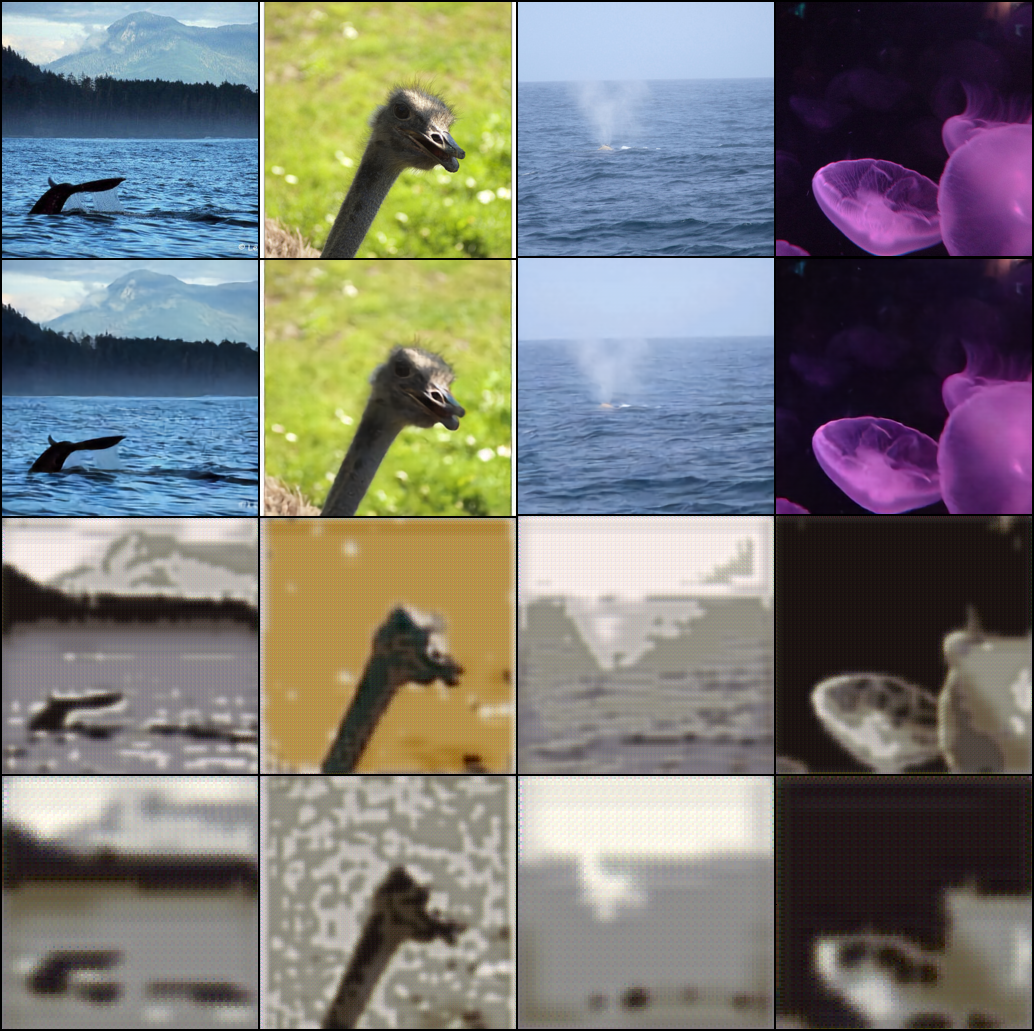}
        \caption{{\small Imagenet}}
        \label{fig:imagenet_recon}
    \end{subfigure}
    \hfill
        \begin{subfigure}[b]{0.24\textwidth}
        \centering
        \includegraphics[width=\textwidth]{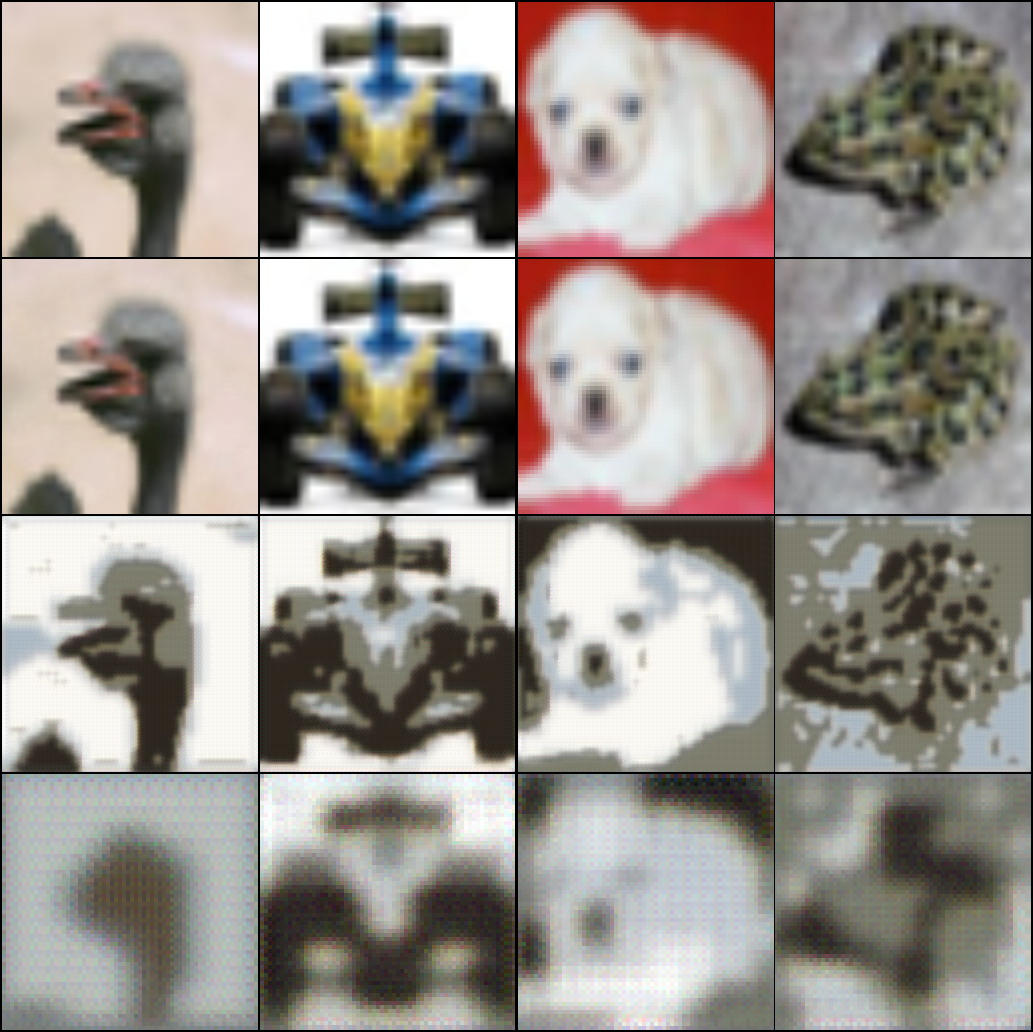}
        \caption{{\small CIFAR10}}
        \label{fig:cifar10_recon}
    \end{subfigure}
    \hfill
    \begin{subfigure}[b]{0.24\textwidth}  
        \centering 
        \includegraphics[width=\textwidth]{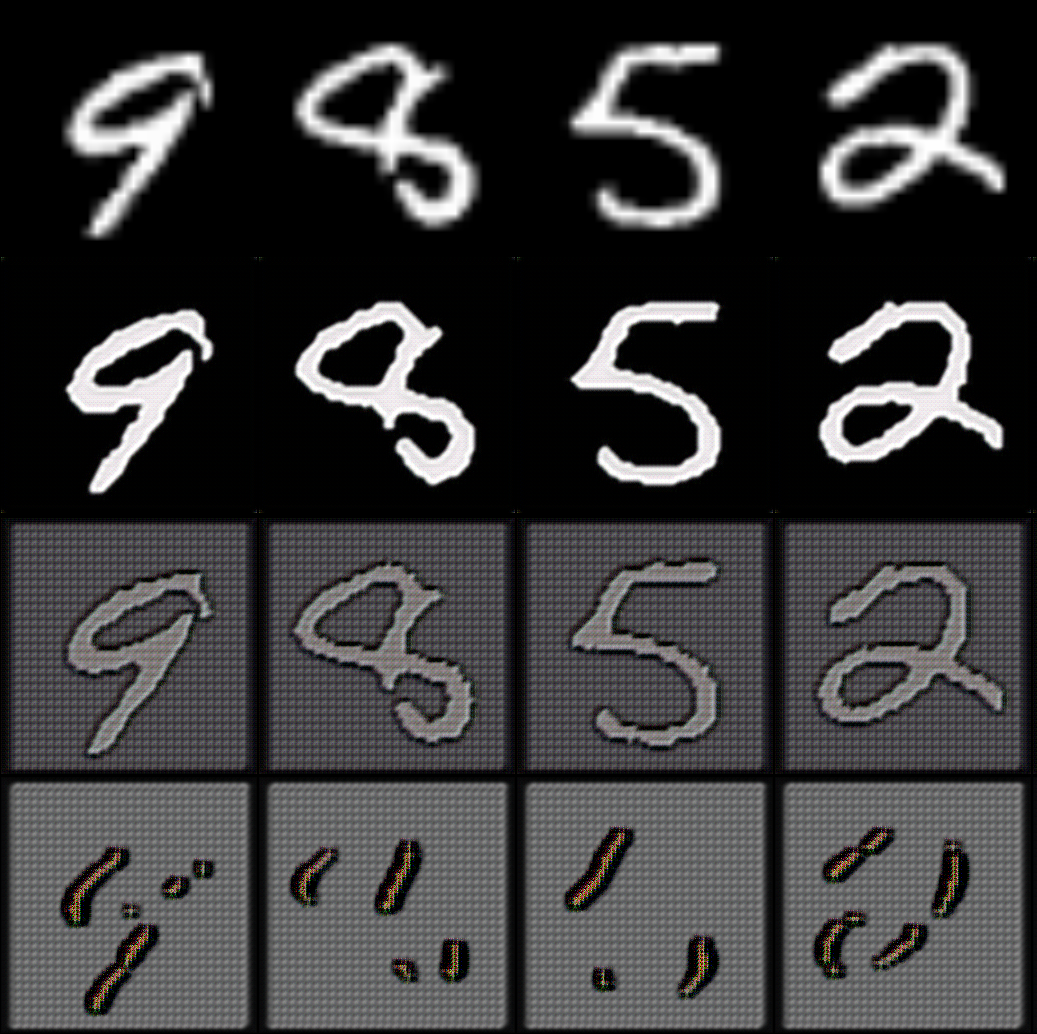}
        \caption{{\small MNIST}}
        \label{fig:mnist_recon}
    \end{subfigure}
    \caption{Reconstructions obtained with HR-VQVAE. First row contains the original images. Second row contains reconstructions using all three layers. Third row indicates reconstructions using the second and third layers. Last row is the reconstructions using only the third layer. Each layer adds extra details to the final reconstruction.}
    \label{fig:multi_comparison}
\end{figure}

%\subsection{Discrete Representation Learning}
%\label{subsec:discrete}
%The development of VAE based techniques for effective discrete representation learning has attracted considerable attention in the research community. 
Rolfe et al.~\cite{rolfe2016discrete} proposed a discrete VAE to train a class of probabilistic models with discrete latent variables. By combining undirected discrete component and a directed hierarchical continuous component, the model efficiently learns both the class of objects in an image and their specific realization in pixels in an unsupervised fashion. Oord et al.~\cite{van2017neural} proposed the vector quantized VAE (VQVAE), a discrete latent VAE model that relies on a vector quantization layer to model discrete latent variables, which quantizes encoder outputs with on-line $k$-means clustering. The discrete latent variables allow the use of a powerful autoregressive model that avoids the posterior collapse problem.
%The key idea of the VQVAE model is to use lossy compression and disregard irrelevant information in reconstructing an image from the latent variables.
Moreover, the model can considerably reduce the amount of information required to reconstruct an image. However, VQVAE suffers from the problem of \textit{codebook collapse}~\cite{dieleman2018challenge}: At some point during training, some portion of the codebook may fall out of use and the model no longer uses the full capacity of the discrete representations,  resulting in a poor reconstruction~\cite{lancucki2020robust}. One of the explanations  of codebook collapse can be found in the typical $k$-means issues \cite{lancucki2020robust} concerning its sensitivity to initialization and non-stationarity of clustered neural activations during training. Moreover,  $k$-means issues  become more severe with the increase of  centroids, and  the ability to encode the input with a broad number of discrete codes decreases~\cite{chorowski2019unsupervised}.
%To achieve compressed images, a vector quantization of the bottleneck representation is obtained by representing the latent variables of the autoencoder with a codebook.
%In \cite{khoshaman2019quantum}, a quantum VAE model was proposed, which discretizes the latent space produced by a VAE using a restricted Boltzmann machine.
%This is a generative model with a classical autoencoding structure and a quantum generative process which is used for quantum and classical data compression.

%Recently, discrete vector quantized autoencoders (VQVAE)~\cite{van2017neural} have shown performance comparable to their stochastic counterparts. %Despite the great efforts in extracting discrete representations for VAEs, most state-of-the-art approaches, e.g., VQVAE, neglect the hierarchical modeling of data, responsible for capturing different levels of abstraction (e.g., low-level details to the high-level context). However, psychology studies suggests that humans concern more about regions of their interest and exclusively want to retrieve images containing relevant parts while ignoring irrelevant image areas (such as the texture regions or background)~\cite{jian2018content}. Therefore, an appropriate hierarchical model can provide hierarchical relations between concepts of interest.
%\hl{VQVAE  uses a flat discrete representation?}.

More recently, several attempts have been made at introducing hierarchical quantized architectures. %, e.g., ~\cite{williams2020hierarchical,takahashi2021hierarchical,razavi2019generating}.
%\subsection{Hierarchical Representation Learning}
%\label{subsec:hierarchical}
%Hierarchical representation learning for generative models has been investigated in the image domain.
In the hierarchical quantized autoencoder~\cite{williams2020hierarchical}, low-resolution discrete representations are decoded to match high-resolution representations and again quantized with a stochastic assignment. For example, 
Takahashi et al.~\cite{takahashi2021hierarchical} proposed a hierarchical representation learning based on VQVAE that enables learning disentangled representations with multiple resolutions independently.
% The method consists of multiple VQVAE modules, each learning the representation with a different resolution.
Razavi et al.~\cite{razavi2019generating} proposed a hierarchical VQVAE, namely VQVAE-2, which extends VQVAE by employing several layers (e.g., top, middle, and bottom layers) of quantized representations to handle hierarchical information in images.
Then, two autoregressive convolutional networks~\cite{albawi2017understanding} were used to model structural and textural information, respectively, to generate new images.
Different layers, however, share the same objective function.
This does not encourage the layers to encode complementary information, and results in inefficient use of the codebooks, as we will show in this paper.
%One of the main disadvantages of VQVAE-2, as we will show in this paper, is that the different quantized layers are not encouraged to encode complementary information from the image because their model shares the same objective function. That leads to redundant representations that do not necessarily correspond to different levels of abstraction and are, therefore, inefficient in representing details on the pixel level. Furthermore, adding more layers results in more time needed for the training process, which makes the model inconvenient for high-load tasks.
Furthermore, VQVAE-2 also suffers from the codebook collapse issue~\cite{zhao2020improve, HCthesis}.

%Besides, the generated textures of the autoregressive network lack fine-grained details due to the lossy nature of VQVAE~\cite{peng2021generating}.

%\begin{figure}
%\vspace{0.5cm}
%\centering
%\includegraphics[width=0.48\textwidth]{figs/orig_vs_recon.png}
%\vspace{-0.3cm}
%\caption{Left: ImageNet $128\times128\times3$ images. Right: reconstructions from a 3-layer %MLVAE with a $32\times32\times1$ latent space. Each layer has 16 codebook indexes.}
%\vspace{-0.5cm}
%\label{fig:orig_vs_recon}
%\end{figure}

In this study, we propose a hierarchical residual learning based vector quantized variational autoencoder (HR-VQVAE) for the image reconstruction and generation tasks.
The first contribution is a novel hierarchical vector quantization encoding scheme.
In contrast with previous research,  our scheme maps the continuous latent representations to several layers of discrete representations through hierarchical codebooks.
Moreover, a novel objective function is proposed to provide contrastive learning by pushing each layer to extract information not learned by its preceding layers. At the same time, the objective optimizes the output image from the combination of representations obtained from all layers (see Fig.~\ref{fig:multi_comparison}).
%As a result, it is not far-fetched that one can expect local information, such as texture, to be reconstructed separately from global information, such as the shape and geometry of objects, generating high-quality and coherent reconstructions (see Fig.~\ref{fig:multi_comparison}).
The hierarchical nature of HR-VQVAE allows us to increase the size of the codebooks without incurring in the codebook collapse problem, resulting in higher quality images. It also provides local access to the codebook layers, thus reducing the search time per layer and speeding up the entire search process. %%%%%%% Should be fit in the Introduction %%%%%%%%%%%
With experiments on well-known image datasets, we show that our model can reconstruct images with higher levels of details and is an order of magnitude faster than state-of-the-art methods (i.e., VQVAE~\cite{rolfe2016discrete} and VQVAE-2~\cite{razavi2019generating}). Moreover, we show that HR-VQVAE can generate high-quality images that challenge some state-of-the-art approaches (i.e., VDVAE~\cite{child2020very} and VQGAN~\cite{esser2021taming}).

The rest of this work is organized as follows. First, we introduce the background in Section~\ref{background}. Then, we present the proposed approach in Section \ref{sec:Proposed-system}. Subsequently, experiments and discussion are given in Section~\ref{sec:experiments}. Finally, we conclude our work in Section \ref{sec:conclusion}.

\begin{figure}
    \centering
    \includegraphics[width=\textwidth]{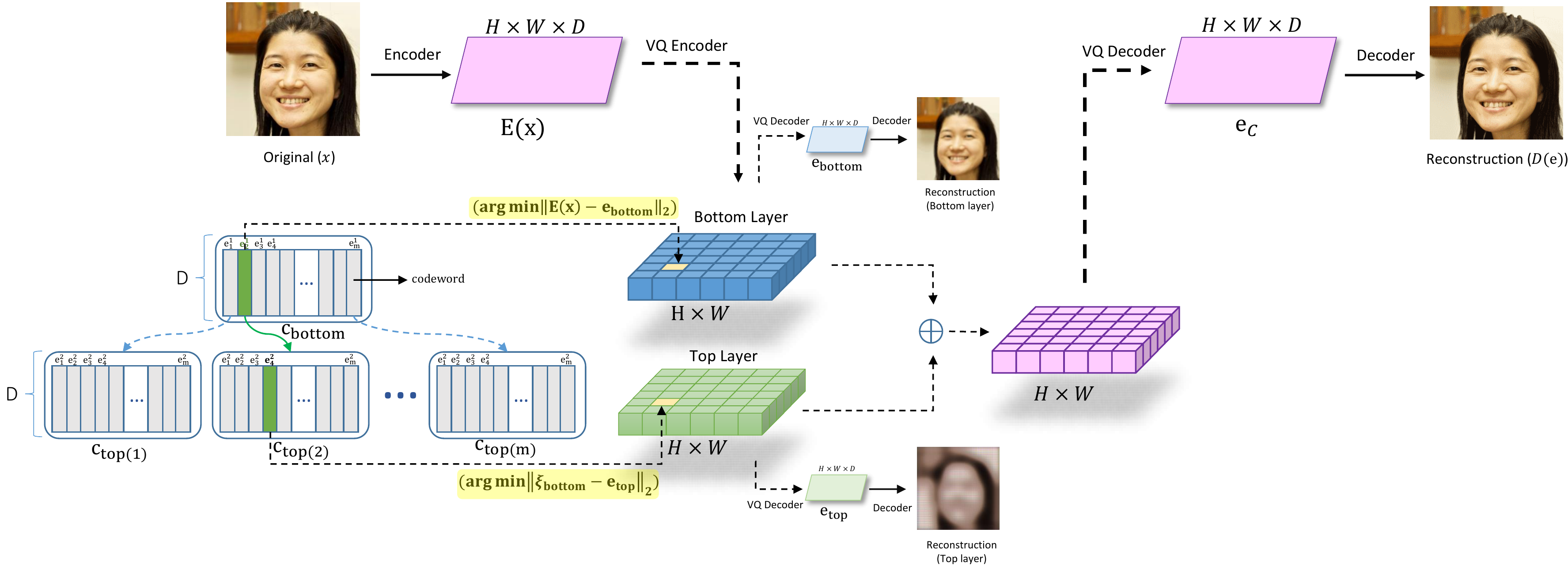}
    \caption{The HR-VQVAE method (only two consecutive layers are shown for simplicity).}
    \label{fig:MLVQVAE_Schematic}
\end{figure}

%%% Background
\section{Background}
\label{background}
\begin{figure}
    \centering
    \includegraphics[width=0.7\textwidth]{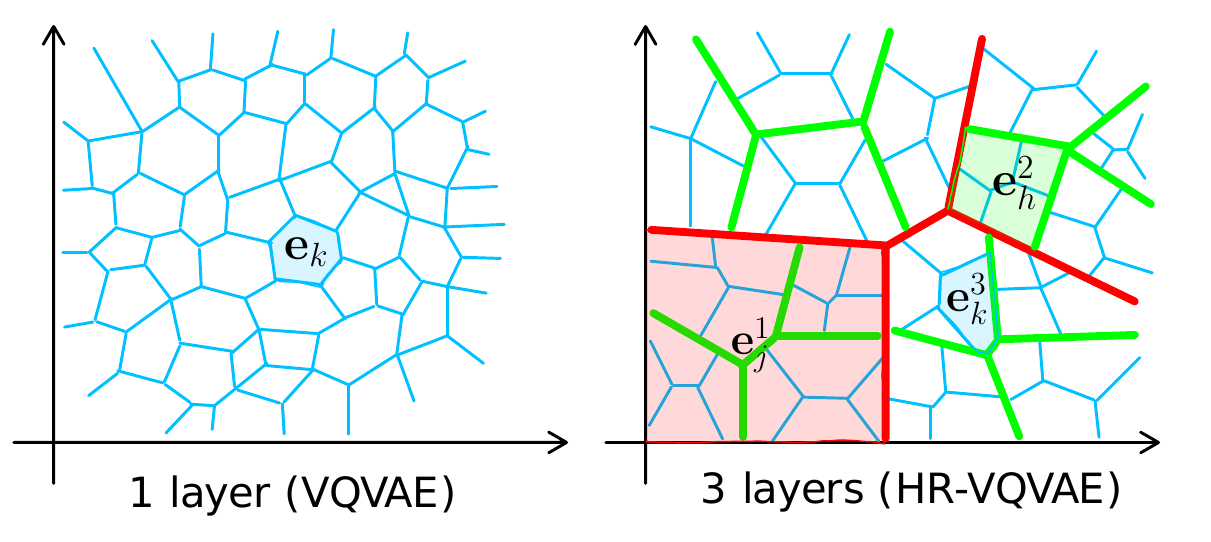}
    \caption{Illustration of vector quantization for 1-layer HR-VQVAE (or VQVAE, left) and 3-layer HR-VQVAE (right). Different colors refer to different layers. For each layer, the  Voronoi cell for one centroid is shaded and annotated as an example (See Eq.~\ref{eq:quantize_vqvae} and \ref{eq:quantize_MLVAE}).}
    \label{fig:VQVAE_vs_MLVAE}
\end{figure}

In this section, we describe aspects of the VQVAE~\cite{van2017neural} and VQVAE-2~\cite{razavi2019generating} models that are necessary to understand the proposed method.
VQVAE first encodes the input image $\mathbf{x} \in \mathbb{R}^\mathsmaller{{H_I\times W_I\times 3}}$ into a continuous latent vector $\mathbf{z} = E(\mathbf{x}) \in \mathbb{R}^\mathsmaller{{H\times W\times D}}$ using a non-linear transformation $E(\cdot)$.
%$E(x)$ is, in general 
Next, each element $\mathbf{z}_{hw} \in \mathbb{R}^\mathsmaller{D}, h\in[1, H], w\in[1, W]$ in the continuous latent representation $\mathbf{z}$ is quantized to the nearest codebook vector (i.e. codeword) $\mathbf{e}_k\in \mathbb{R}^\mathsmaller{D},~k \in 1,...,m$ by
\begin{equation}
\label{eq:quantize_vqvae}
    \text{Quantize}(\mathbf{z}_{hw}) = \mathbf{e}_k \text{ where } k=\arg \min_{j} ||\mathbf{z}_{hw}-\mathbf{e}_j||_2,
\end{equation}
as illustrated in Fig.~\ref{fig:VQVAE_vs_MLVAE} (left).
The quantized vectors corresponding to each element $\mathbf{z}_{hw}$ are then recombined into the quantized representation $\mathbf{e} \in \mathbb{R}^\mathsmaller{{H\times W\times D}}$ to form the input to a decoder that reconstructs the original image through a non-linear function $\mathcal{D(\cdot)}$.
%We call $\mathbf{z}$ the discrete latent variable that holds the 
%To learn these mappings, the gradient of the reconstruction error is then back-propagated through the decoder, and to the encoder using the straight-through gradient estimator.
The encoder $E(\cdot)$, the codeword $\{\mathbf{e}_k\}$, and the decoder $\mathcal{D}(\cdot)$ are learned from data by optimizing the objective function
%\begin{multline}
%\label{eq:loss_vqvae}
%\mathcal{L}(\mathbf{x},\mathcal{D}(\mathbf{e})) = \lVert\mathbf{x}-\mathcal{D}(\mathbf{e})\rVert^2_2 + \lVert\text{sg}[\mathbf{z}] - \mathbf{e}\rVert^2_2 + \\
%    +\beta\lVert\text{sg}[\mathbf{e}] - \mathbf{z}\rVert^2_2.
%\end{multline}
\begin{equation}
\label{eq:loss_vqvae}
\mathcal{L}(\mathbf{x},\mathcal{D}(\mathbf{e})) = \lVert\mathbf{x}-\mathcal{D}(\mathbf{e})\rVert^2_2 + \lVert\text{sg}[\mathbf{z}] - \mathbf{e}\rVert^2_2 + \beta\lVert\text{sg}[\mathbf{e}] - \mathbf{z}\rVert^2_2.
\end{equation}
This function aims at minimizing the reconstruction error $\lVert\mathbf{x}-\mathcal{D}(\mathbf{e})\rVert_2$ whilst minimizing the quantization error $\lVert\mathbf{z}-\mathbf{e}\rVert_2$.
In Eq.~\ref{eq:loss_vqvae}, $\text{sg}(.)$ refers to a stop-gradient operator that cuts the gradient flow through its argument during the backpropagation, and $\beta$ is a hyperparameter which controls the reluctance to change the latent representation corresponding to the encoder output.
%%%%%%%%%%%%%%%%%%%%%%%%%%%%%%%

% VQVAE, the prior distribution over the discrete latent variable is a categorical distribution, which is kept constant and uniform during training. %After training, the prior is made autoregressive to allow sample generation via ancestral sampling. The authors used PixelCNN over the discrete latents for images, and a WaveNet for raw audio. 

%In the training phase, VQVAE fits an autoregressive distribution over discrete latent representations %$z$  and a categorical prior distribution over discrete latent representations.
%Given the input data $\mathbf{x}$ and discrete latent variables $\mathbf{z}$, because the prior distribution over the discrete latent representations is a categorical distribution, and can be made autoregressive by depending on other $\mathbf{z}$ in the latent space, VQVAE fits prior over the discrete representation space using PixelCNN to leverage of autoregressive neural networks as strong density estimators and at the same time has the right hierarchical structure for representation learning.
%%%%%%%%%%%%%%%%%%%%%%%%%%%%%%

%A follow-up to the VQVAE model, named VQVAE-2. % (see Fig.~\ref{fig:VQVAE2_Schematic}).
VQVAE-2 extends VQVAE to attain a hierarchy of vector quantized codes. It compresses images into several latent spaces, from the \textit{top} layer (smaller size) to the \textit{bottom} layer (larger size), which is conditioned on the top layer in order for the top layer to extract general information from the image and the bottom layer to add more detail in the image reconstruction. The codebooks at different layers, however, are not related by a hierarchy.

\section{Proposed Approach}
\label{sec:Proposed-system}
%%% This part can be rewritten to give the general reason for such a modeling %%%%%%%%%%%%%%%%%%%%%%%%%%%%
The architecture of the proposed HR-VQVAE is illustrated in Fig.~\ref{fig:MLVQVAE_Schematic}, where we only show two consecutive layers for simplicity.
%MLVAE addresses the shortcomings of the state-of-the-art latent space extraction techniques, VQVAE and VQVAE-2, by introducing a hierarchy of quantized representations that are encouraged to learn different levels of abstractions in the data.
%%%%%%%%%%%%%%%%%%%%%%%%%%%%%%%%%%%%%%%%%%%%%%%%%%%%%%
%employing the idea of quantization technique~\cite{van2017neural,gersho2012vector} to capture different hierarchical layers of abstraction.
%MLVAE comprises two consecutive phases:
%\begin{enumerate}
%    \item Creating hierarchical pre-trained latent codebooks obtained using samples in the training dataset.
%    \item Fitting a prior to the latent codebooks.
%\end{enumerate}
%MLVAE comprises two consecutive steps as follows.\\
%%%%%%%%%%%%%%%%%%%%%%%%%%%%%%%%%%%%%%%%%
%\subsection{Phase 1: Hierarchical Latent Codebooks}
As in VQVAE, the original image $\mathbf{x}$ is first encoded into continuous embeddings that we call $\boldsymbol{\xi}^0 = E(\mathbf{x})$ by a non-linear encoder.
Differently from VQVAE, however, these embeddings are then iteratively quantized into $n$ hierarchical layers of discrete latent variables.
Assuming the first layer has a codebook of size $m$, the second layer will have $m$ codebooks of size $m$, and so on for subsequent layers.
In general, layer $i$ has $m^{i-1}$ codebooks of size $m$, for a total of $m^i$ codewords.
However, only one of those codebooks is used in each layer depending on which codewords where chosen in the previous layers.
This is illustrated in Fig.~\ref{fig:MLVQVAE_Schematic} where the vector selected within $C_\text{bottom}$ determines the codebook that is activated in the top layer (in this case $C_\text{top}(2)$).
Such a hierarchical searching procedure provides the advantage of local access to codebook indexes, which dramatically reduces search time.
Fig.~\ref{fig:VQVAE_vs_MLVAE} (right) exemplifies this structure where the number of layers $n=3$ and the codebook size $m=4$.
The resulting Voronoi cells are shown in red, green and blue for the first, second and third layer, respectively.

In each layer $i$, the codebook is optimized to minimize the error between the codewords $\mathbf{e}^i_k \in \mathbb{R}^\mathsmaller{D}$ and the elements $\boldsymbol{\xi}^{i-1}_{hw} \in \mathbb{R}^\mathsmaller{D}$ of the residual error from the previous layer:
\begin{equation}
\label{eq:quantize_MLVAE}
    \text{Quantize}^i(\boldsymbol{\xi}^{i-1}_{hw}) = \mathbf{e}^i_k \text{ where } k=\arg \min_{j} \lVert\boldsymbol{\xi}^{i-1}_{hw}-\mathbf{e}^i_j\rVert_2,
\end{equation}
and $\mathbf{e}_k^i$ belongs to one of the possible codebooks $C_{i}(t)$ for layer $i$.
Which codebook is used is determined by the codeword $\mathbf{e}_t^{i-1}$ selected at the previous layer. 

%Therefore, for layer $i$, $\boldsymbol{\xi}^{i-1}$ is quantized  based on its distance to the prototype vectors in the codebook of $i^{th}$ layer, such that the vector $\boldsymbol{\xi}^{i-1}$ is replaced by the index of the nearest prototype vectors in the codebook $\mathbf{e}^i_k,~k \in (1,...,m)^{C_{i(t)}}$ and is transmitted to the VQ decoder,
%where $t$ is the selected index in the {\scriptsize$(i-1)^{th}$} layer codebook, defining which codebook in the layer $i$ (i.e. $C_{i(t)}$) must be selected for the searching process.
%Note that Fig.~\ref{fig:MLVQVAE_Schematic} only shows two consecutive layers called ``Bottom'' and ``Top'' for simplicity.
%where $\mathbf{e}^i$ is the quantized representations of layer $i$, and $\boldsymbol{\xi^{i-1}}$ is a parameter that contains the error which is what has not been learned from the latent space by the first to {\scriptsize$(i-1)^{th}$} layers, therefore pushing the layer $i$ to learn different information from previous layers, $t$ is the selected index in the {\scriptsize$(i-1)^{th}$} layer codebook which define which codebook in the layer $i$ (i.e. $C_{i(t)}$) must be selected for the searching process.
Within each layer, the codewords $\mathbf{e}^i_k$ are combined to form the tensor $\mathbf{e}^i \in \mathbb{R}^\mathsmaller{{H\times W\times D}}$.
Across the different layers, we then combine the tensors $\mathbf{e}^i$  to form the ``combined'' discrete representation $\mathbf{e}_C$ which, in turn, is fed into the decoder that reconstructs the image $\mathbf{x}$.
%This combination can be linear with fixed weights, or mediated by a set of parameters $\boldsymbol{\alpha}^i \in\mathbb{R}^\mathsmaller{{H\times W\times D}}$ that are also learned from the data:
\begin{equation}
    \label{eq:codebook_sum}
    \mathbf{e}_C = \sum_{i=1}^{n} \mathbf{e}^i,
    %\mathbf{e}_C = (\boldsymbol{\alpha}^i)^T \mathbf{e}^i,
\end{equation}
%with $\sum_{i=1}^n{\boldsymbol{\alpha}^i} = \mathbf{1}_{H\times W\times D}$, the one tensor of size $H\times W\times D$, and $\odot$ is the element-wise multiplication.
%The parameters $\boldsymbol{\alpha}^i$ weight the contributions of each hierarchical codeword $\mathbf{e}^i$ at the latent pixel level.
By doing this, we allow the combined discrete latent representation $\mathbf{e}_C$ to incorporate different aspects of the image, depending on the area that we try to reconstruct.
%In the experiments we investigate the effect of plain linear combination of $\mathbf{e}^i$ as opposed to combinations using $\boldsymbol{\alpha}^i$.
%therefore focus on different aspects and different levels of abstaction.
%By giving more weight to more important pixels in different quantized layers, we increase the flexibility of our model and allow it to.
%Next, the VQ decoder first remaps the received indices of layers to their corresponding vectors in their codebooks (i.e. $\mathbf{e}^1,...,\mathbf{e}^n$) and then combines them together to make the main mixed vectors (i.e. $\mathbf{e}$).  Such a combination can be done using the hyperparameter $\alpha$ (see Eq.~\ref{eq:codebook_sum}). Afterward, the main non-linear decoder is fed by the output of the VQ decoder and reconstructs the input sample $\mathbf{x}$. The whole objective function of MLVAE is as follows,
The objective function used to train the system is:
%\begin{multline}
%\label{eq:loss_mlvae}
%\mathcal{L}({\mathbf{x}},\mathcal{D}({\mathbf{e}_C})) = \lVert{\mathbf{x}}-\mathcal{D}({\mathbf{e}_C})\rVert^2_2 + \lVert\text{sg}[\boldsymbol{\xi}^0] - {\mathbf{e}_C}\rVert^2_2 + \\
%    +\beta_0\lVert\text{sg}[{\mathbf{e}_C}] - \boldsymbol{\xi}^0\rVert^2_2 +\sum_{i=1}^{n}\mathcal{L}(\boldsymbol{\xi}^{i-1}, \mathbf{e^i}),
%\end{multline}
\begin{equation}
\label{eq:loss_mlvae}
\mathcal{L}({\mathbf{x}},\mathcal{D}({\mathbf{e}_C})) = \lVert{\mathbf{x}}-\mathcal{D}({\mathbf{e}_C})\rVert^2_2 + \lVert\text{sg}[\boldsymbol{\xi}^0] - {\mathbf{e}_C}\rVert^2_2 
    +\beta_0\lVert\text{sg}[{\mathbf{e}_C}] - \boldsymbol{\xi}^0\rVert^2_2 +\sum_{i=1}^{n}\mathcal{L}(\boldsymbol{\xi}^{i-1}, \mathbf{e^i}),
\end{equation}
with
\begin{equation}
    \label{eq:loss_layer}
    \mathcal{L}(\boldsymbol{\xi}^{i-1}, \mathbf{e^i}) = \lVert\text{sg}[\boldsymbol{\xi} ^{i-1}] - \mathbf{e}^i\rVert^2_2 
    +\beta_i \lVert\text{sg}[\mathbf{e}^i] - \boldsymbol{\xi}^{i-1}\rVert^2_2,
\end{equation}\\
%where $\mathbf{e}$ is the main quantized code obtained from Eq.~\ref{eq:codebook_sum} for the training example $\mathbf{x}$,
where $\beta_i$ are hyperparameters which control the reluctance to change the code corresponding to the encoder output.
%%%%%%%%%%%%%%%% Must be rewritten. Very badly written %%%%%%%%%%%%%%%%%%%%

The main goal of Eqs.~\ref{eq:loss_mlvae}, and \ref{eq:loss_layer} is to make a hierarchical mapping of input data in which each layer of quantization extracts residual concepts from its bottom layers. In this regard, $\boldsymbol{\xi}^i$ (Eq.~\ref{eq:loss_layer}) plays an essential role in making the hierarchically learning of layers which makes the main differences between our model and the VQVAE-2 model.
%As mentioned before, we train a learnable hyperparameters, $\boldsymbol{\alpha}^i$, which learns to give more weight to more important pixels in different quantized layers. Such hyperparameters increase the flexibility of our model in the combination of quantized vectors.% (i.e. Eq.~{\ref{eq:codebook_sum}}).
It should be noted that both VQ encoder and decoder share the same hierarchical codebooks.

Finally, as in VQVAE, for each $\mathbf{e_ C}$ we fit a prior distribution to all training samples using an autoregressive model (PixelCNN~\cite{van2016conditional}).
Such a model factorizes the joint probability distribution over the input space into a product of conditional distributions for each dimension of the sample.
%with respect to all the previous pixels in a predefined order (e.g., raster scan ordering): $p_\theta(\mathbf{x}) = \prod_{i=0}^np_\theta(x_i|\mathbf{x}_{<i})$.
For generation of new images we use ancestral sampling taking advantage of the chain rule of probability.

\begin{figure}
    \centering
    \includegraphics[width=0.8\textwidth]{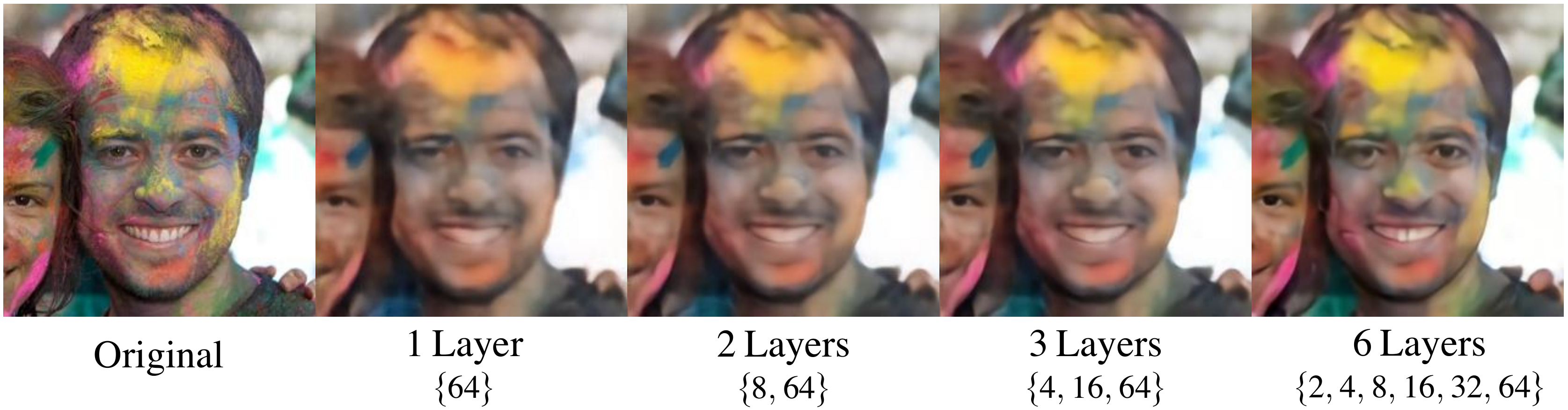}
    \caption{Reconstructions obtained with HR-VQVAE models with different depths (i.e., number of layers). The latent maps are $32\times32$, and the number of codewords for each layer is specified from bottom to top in order from right to left for each model.}
    %the same: $64^1$, $8^2$, $4^3$ and $2^6$ (from left to right).}
    \label{fig:MLVQVAE_FFHQ}
\end{figure}

\section{Experiments and Discussion}
\label{sec:experiments}
We conducted our experiments on four well-known datasets, FFHQ~\cite{karras2019style} ($256\times256$), ImageNet~\cite{deng2009imagenet} ($128\times128$),   CIFAR10~\cite{krizhevsky2009learning} ($32\times32$) and MNIST~\cite{lecun1998gradient} ($28\times28$).
We start this section by investigating the effect of varying the depth of the hierarchy in our model.
%%%%%% Should be rewritten in a better way %%%%%%%%%%%%%%%%%%
To this end, we defined models with $n$ layers and $m$ codewords per codebook.
As explained in Sec.~\ref{sec:Proposed-system}, the number of codewords in each layer $i$ is $m^i$, and, therefore the layers will have $\{m, m^2, \dots, m^n\}$ codewords.
To ensure the same level of resolution among the models we compare models with the same number of codewords in the final layer, which corresponds to the maximum resolution. %%%%%%%%%%%%%%%%%%%%%%%%%%%%%%%%%%%%%%%%%%%%%%%%%%%%%%%%%%%%%
Fig.~\ref{fig:MLVQVAE_FFHQ}, shows HR-VQVAE reconstructions with different numbers of layers, namely from one to six.
%\hl{The total number of codewords in the last layer serves as an upper bound for the degree of the freedom of the model. Therefore, for fair comparison, all of the model configurations contain the same amount of codewords in the last layer.} 
Although all configurations have 64 codewords in the final layer, we observe that increasing the depth of the model results in reconstructions with more details (zoom into the pdf version).
A possible explanation for such an improvement is that the hierarchical nature of the codebooks acts as regularization during training and allows the model to allocate codewords more efficiently.
%Note that HR-VQVAE-\{64\} is equal to VQVAE with 64 codewords in the codebook.
%\hl{In addition, the figure shows that when we increase the number of layers the model can reach a good reconstruction using smaller codebooks, resulting in higher compression ratio.}

\begin{figure}
    \centering
    \includegraphics[width=0.8\textwidth]{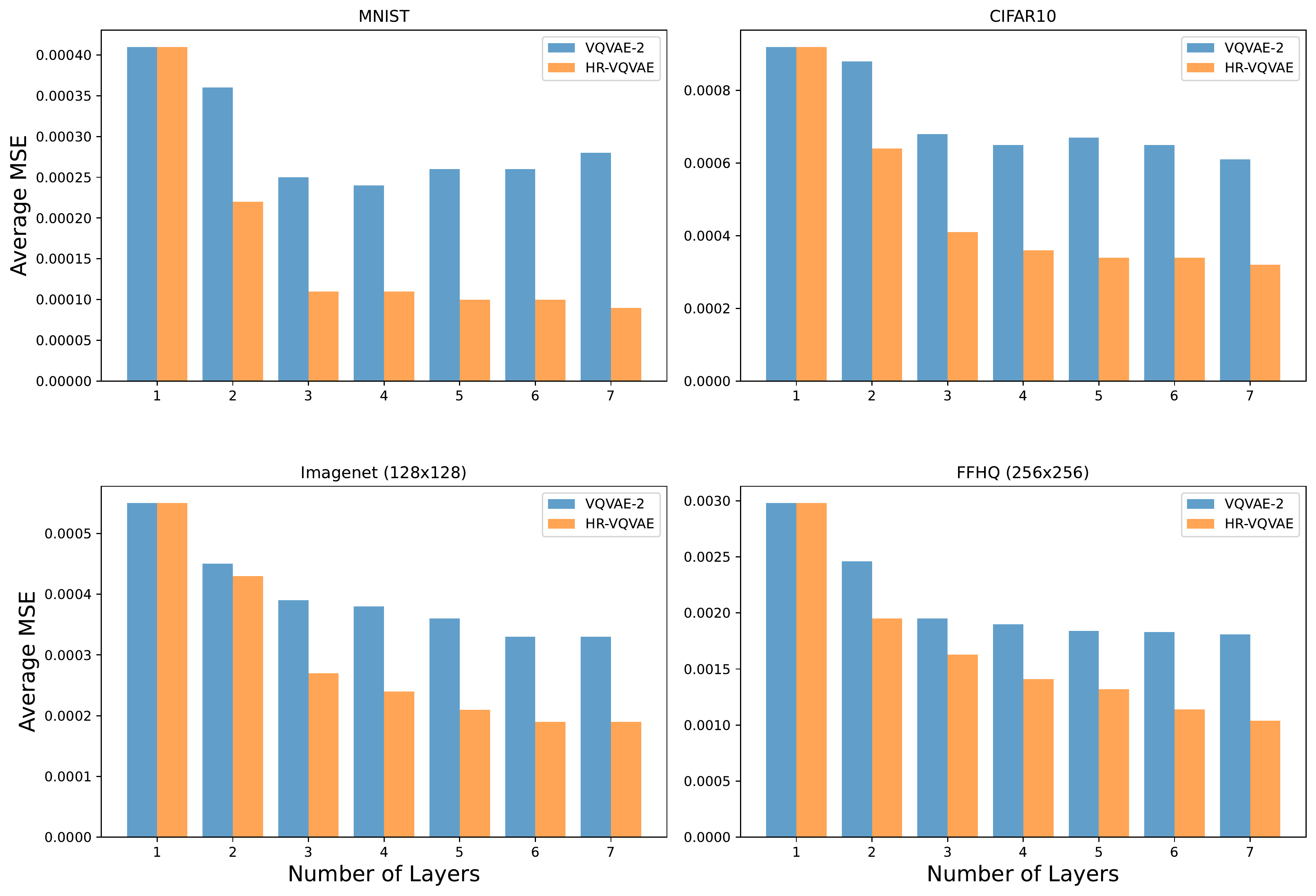}
    \caption{The effect of model depth (number of layers) on image reconstruction.}
    \label{fig:layer_nums_recons}
\end{figure}

Fig.~\ref{fig:layer_nums_recons} provides a comparison with VQVAE-2 on the effect of the model depth (i.e., number of layers) in terms of the reconstruction mean square error (MSE)~\cite{zhang2005color}. The results demonstrate that increasing the model depth leads to better performance of HR-VQVAE compared to VQVAE-2. Furthermore, the performance of HR-VQVAE improves consistently for all datasets with the increase in the number of layers. However, increasing the number of layers does not improve the performance of VQVAE-2 (for Imagenet and FFHQ) from a certain point, and in some cases (MNIST and CIFAR10), the performance decreases. In the following experiments, we will use three layers in HR-VQVAE to be able to compare with VQVAE-2 which also uses three layers, while VQVAE uses a single layer.

\begin{figure}
    \centering
    \includegraphics[width=0.8\textwidth]{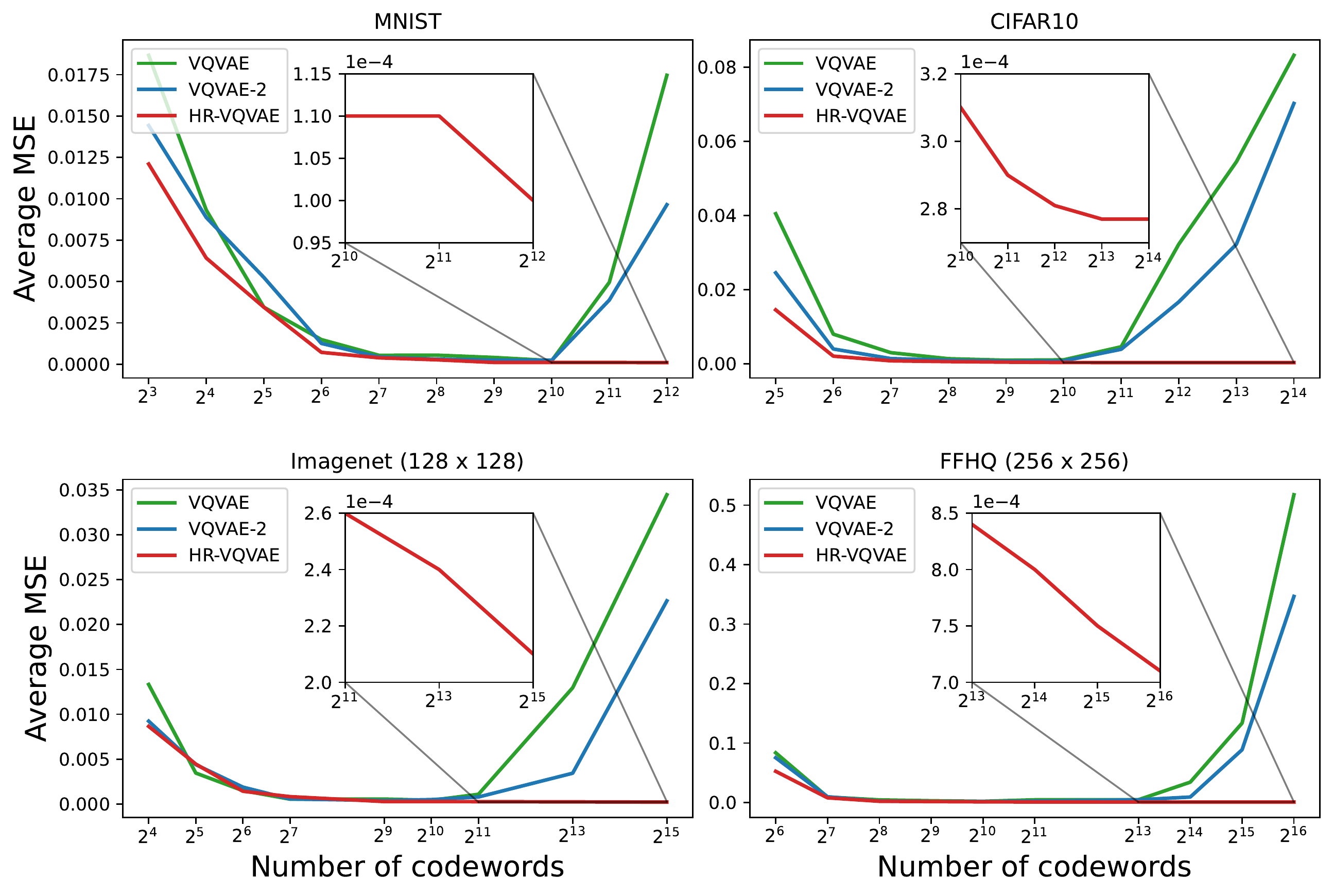}
    \caption{Average MSE vs number of codewords for different datasets and methods. Both VQVAE and VQVAE-2 collapse when the codebook size is increased over a certain limit. However, HR-VQVAE continues improving as shown in the zoom windows inside each plot.}
    \label{fig:codebook_collapse}
\end{figure}

\begin{figure*}
    \centering
    \includegraphics[width=\textwidth]{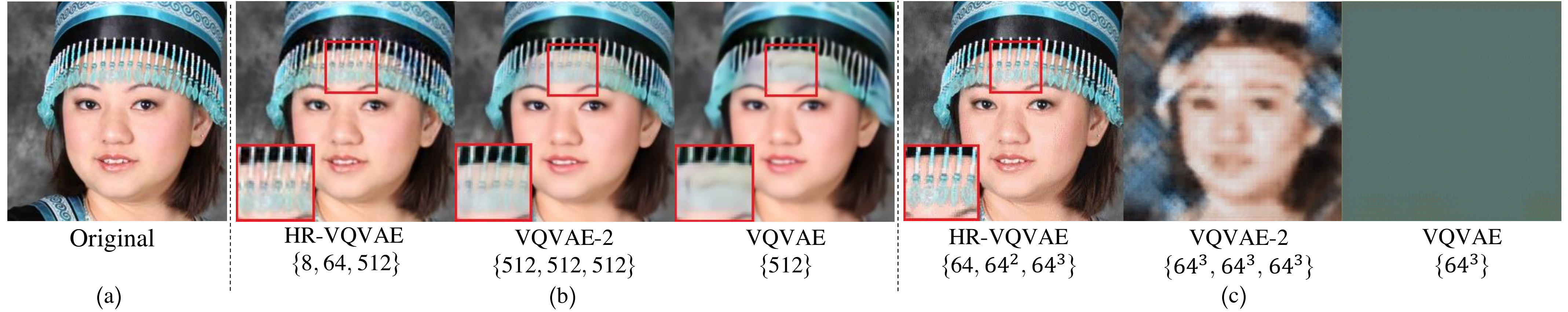}
    \caption{Reconstructions obtained with HR-VQVAE, VQVAE-2 and VQVAE. Number of codewords per layer are given for each model. Both VQVAE and VQVAE-2 are clearly affected by the codebook collapse problem.}
    \label{fig:mlvae_vs_recons}
\end{figure*}

We first compare the effect of increasing the codebook size in our model as well as VQVAE and VQVAE-2. Fig.~\ref{fig:codebook_collapse} illustrates the behavior of HR-VQVAE and the baseline models with different numbers of codewords. As it can be seen, by increasing the number of codewords up to a certain number, the performance of all models improves, whereas HR-VQVAE shows higher performance. However, the efficiency of the baseline models starts decreasing from a certain point with increasing the number of codewords, while the efficiency of HR-VQVAE continuously increases for all datasets. This means that not only HR-VQVAE does not suffer from the codebook collapse problem, but it can also benefit from increasing the number of codebooks to improve performance.
Fig.~\ref{fig:mlvae_vs_recons} provides a visual example for Fig.~\ref{fig:codebook_collapse}. Fig.~\ref{fig:mlvae_vs_recons} (b) shows reconstructions where the size of codebooks is 512 for VQVAE, $\{512, 512, 512\}$ for VQVAE-2 and $\{8, 64, 512\}$ for HR-VQVAE.
Similarly to Fig.~\ref{fig:MLVQVAE_FFHQ}, HR-VQVAE produces superior details than VQVAE with the same codebook size (zoom into the pdf version).
VQVAE-2 produces a very smooth image but misses some of the details.
More interesting is to study what happens if we increase the codebook size in all the models.
Fig.~\ref{fig:mlvae_vs_recons} (c) shows that both VQVAE and VQVAE-2 are affected by codebook collapse.
On the contrary, HR-VQVAE can take full advantage of the increased complexity and produces the best reconstruction of this list.

\begin{figure}
    \centering
    \includegraphics[width=0.6\columnwidth]{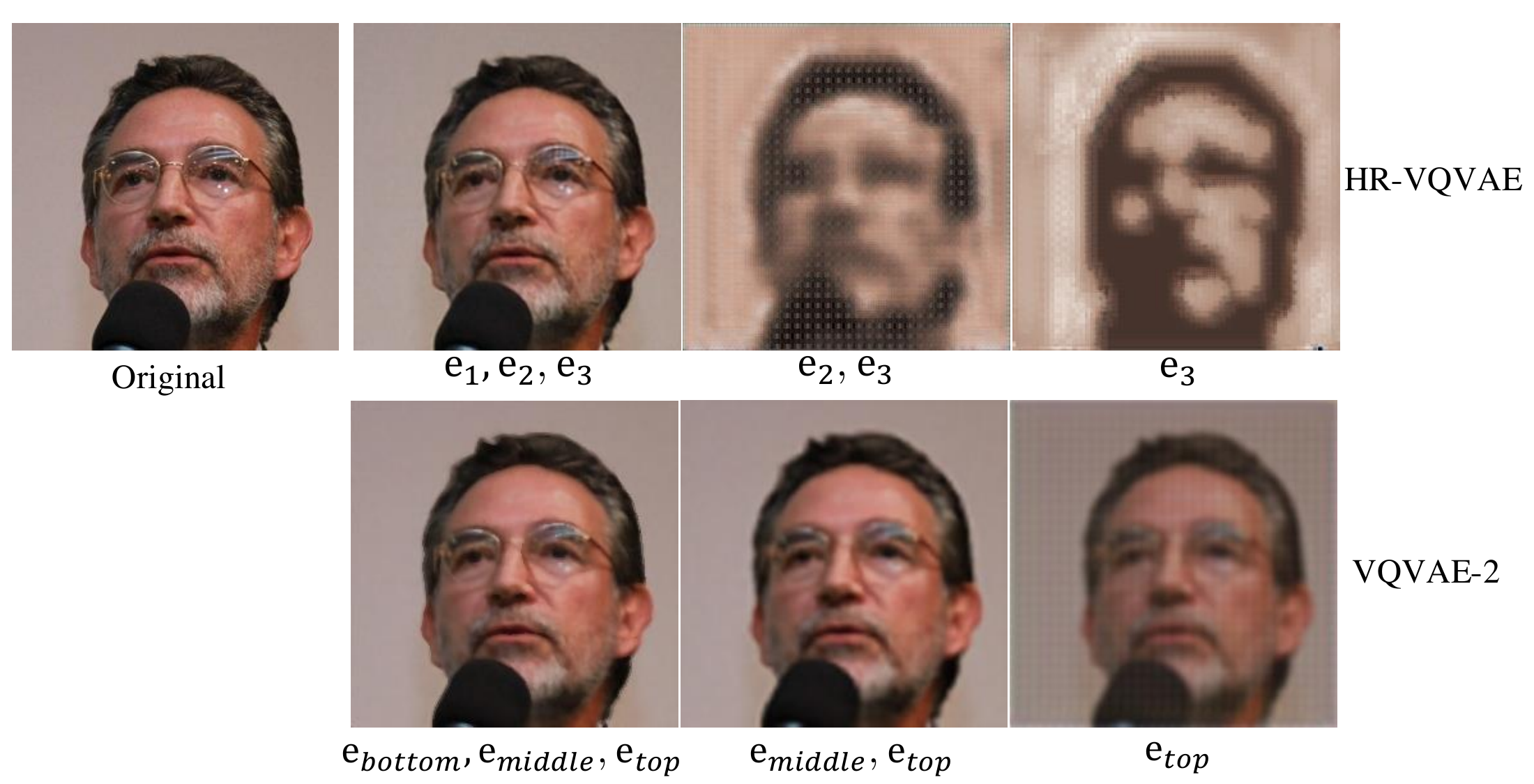}
    \caption{Reconstruction comparison of HR-VQVAE and VQVAE-2.}
    \label{fig:MLVAE_vs_VQVAE-2}
\end{figure}

Fig.~\ref{fig:MLVAE_vs_VQVAE-2} compares 3-layers HR-VQVAE and 3-layer VQVAE-2 to illustrate the different information encoded in different layers in the two models.
HR-VQVAE image reconstructions (first row) attain a better reconstruction quality with  more details than VQVAE-2 (second row).
One possible explanation is that HR-VQVAE encourages the different layers to encode different information about the image; whereas the information in VQVAE-2 is strongly overlapping.
This may result in a less efficient latent representation.

Table~\ref{tbl:reconstruction_error} reports the mean squared error (MSE) and fréchet inception distance (FID)~\cite{lucic2018gans} results for HR-VQVAE, VQVAE, VQVAE-2 for image reconstructions.
The reported scores confirm all the results presented so far.
Our proposed HR-VQVAE is able to outperform the baseline models for image reconstructions on all datasets in terms of both MSE and FID score, which is further evidence of the efficiency of our model.

\begin{table}
    \centering
    \begin{tabular}{lcccc}
    \hline\hline
    \multirow{2}{*}{Model} & \multicolumn{4}{c}{FID $\downarrow$ / MSE $\downarrow$}\\
    & FFHQ & ImageNet & CIFAR10 & MNIST\\
    \hline
    VQVAE~\cite{rolfe2016discrete}& 2.86/0.00298 & 3.66/0.00055 & 21.65/0.00092 & 7.9/0.00041\\
    VQVAE-2~\cite{razavi2019generating} & 1.92/0.00195 & 2.94/0.00039 & \textbf{18.03}/0.00068 & 6.7/0.00025\\
    HR-VQVAE & \textbf{1.26/0.00163} & \textbf{2.28/0.00027} & 18.11/\textbf{0.00041} & \textbf{6.1/0.00011}\\
    \hline\hline
    \end{tabular}
    \caption {FID/MSE reconstruction results using HR-VQVAE, VQVAE-2 and VQVAE.}
    \label{tbl:reconstruction_error}
\end{table}

%Furthermore, it can improve the training procedure in terms of number of iterations required
As mentioned in the introduction, the hierarchical structure of the codebooks in HR-VQVAE provides fast access to codebook indexes across layers which significantly reduces the search time during decoding. Table~\ref{tbl:search_time} reports a comparison of execution time for the high-quality reconstructions of 10000 samples for HR-VQVAE as well as VQVAE and VQVAE-2.
%The inputs to the models are 256$\times$256 and 128$\times$128, 32$\times$32 and 28$\times$28 images for the  FFHQ, Imagenet, CIFAR10 and MNIST datasets, respectively.
The input images are compressed to quantized latent codes of size 32$\times$32 for FFHQ and Imagenet and 16$\times$16 for CIFAR10 and MNIST in HR-VQVAE and VQVAE.
For the VQVAE-2 model, the images are compressed into latent codes of size \{32$\times$32, 16$\times$16, 8$\times$8\} for the bottom, middle, and top layers, respectively for FFHQ and Imagenet and \{16$\times$16, 8$\times$8, 4$\times$4\}, respectively for CIFAR10 and MNIST.
Table~\ref{tbl:search_time} reports that HR-VQVAE reaches an over ten-fold increase in reconstruction speed compared to VQVAE-2, and a large improvement with respect to VQVAE.
Although HR-VQVAE has codebook sizes of $\{ m, m^2, \dots, m^n \}$ in the different layers, it only needs to search through $n\times m$ such vectors due to its hierarchical structure.

\begin{table}
    \centering
    \begin{tabular}{lcccc}
    \hline\hline
    \multirow{2}{*}{Model} & \multicolumn{4}{c}{Seconds}\\
    & FFHQ & Imagenet & CIFAR10 & MNIST\\
    \hline
    VQVAE~\cite{rolfe2016discrete} & 5.0977652 & 4.6152677 & 2.7087896 & 0.062474\\
    VQVAE-2~\cite{razavi2019generating} & 9.3443758 & 8.8135872 & 4.4492340 & 0.090778\\
    HR-VQVAE & \textbf{0.8398101} & \textbf{0.6714823} & \textbf{0.4667842} & \textbf{0.010830}\\
    \hline\hline
    \end{tabular}
    \caption{Time for reconstructing 10000 samples using HR-VQVAE, VQVAE-2 and VQVAE.}
    \label{tbl:search_time}
\end{table}

\begin{figure}
    \centering
    \raisebox{3.5em}{\rotatebox{90}{HR-VQVAE}}
    \begin{subfigure}[b]{0.23\textwidth}
        \centering
        \includegraphics[width=\textwidth]{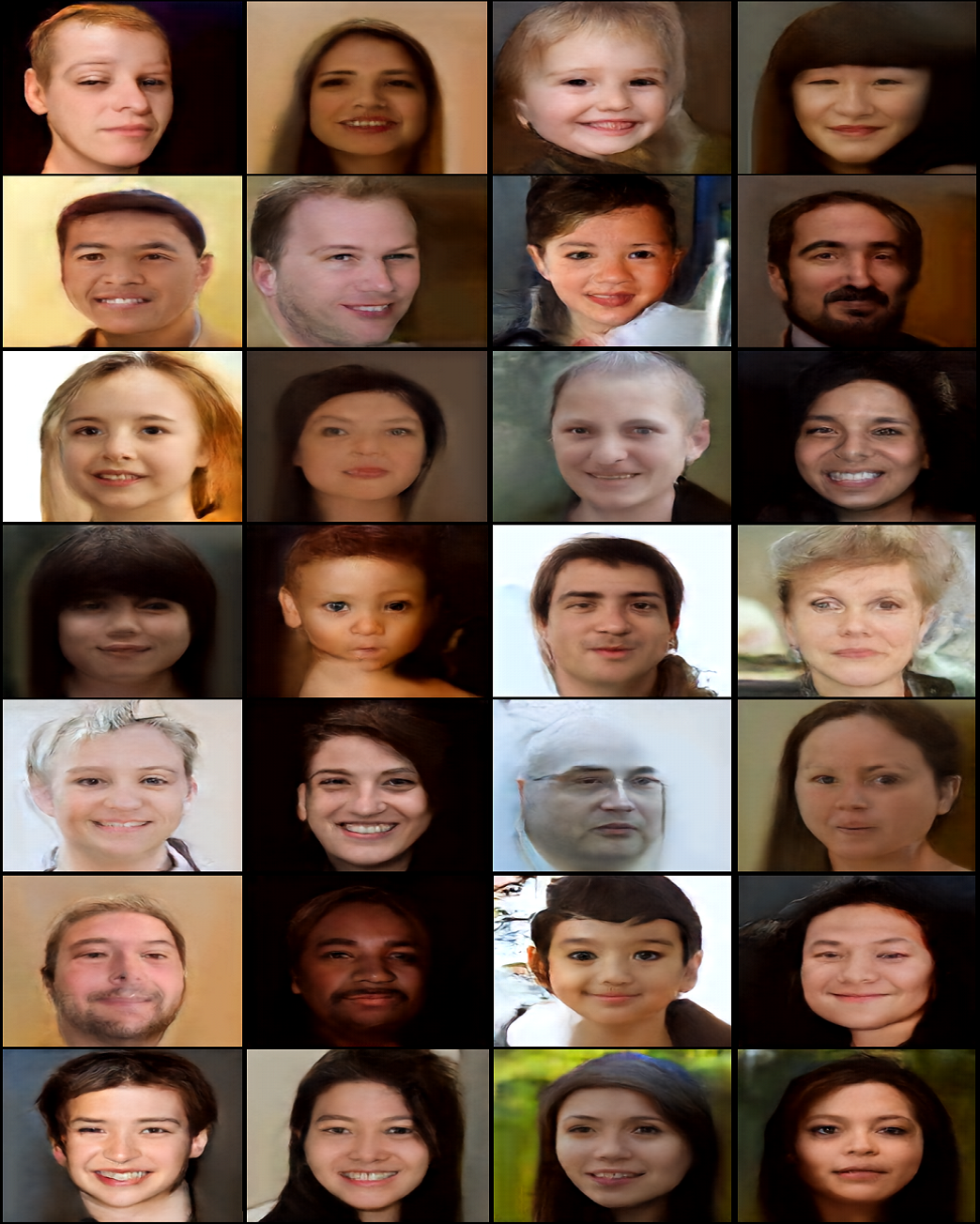}
        \label{fig:ffhq_gen}
    \end{subfigure}
    \hfill
    \begin{subfigure}[b]{0.23\textwidth}  
        \centering 
        \includegraphics[width=\textwidth]{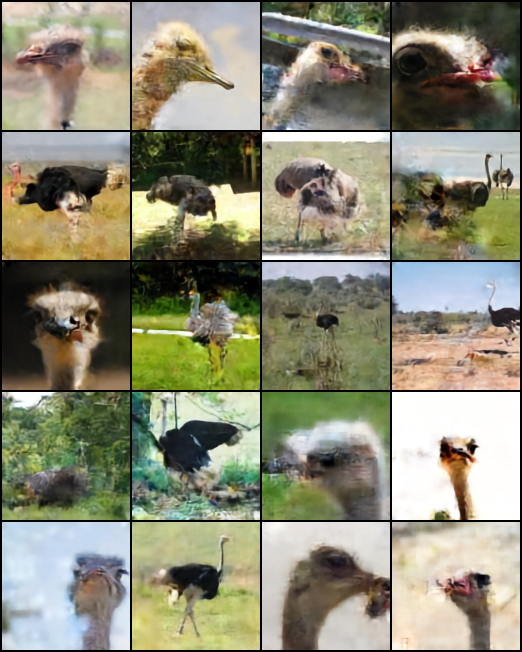}
        \label{fig:imagenet_gen}
    \end{subfigure}
    \hfill
    \begin{subfigure}[b]{0.23\textwidth}
        \centering
        \includegraphics[width=\textwidth]{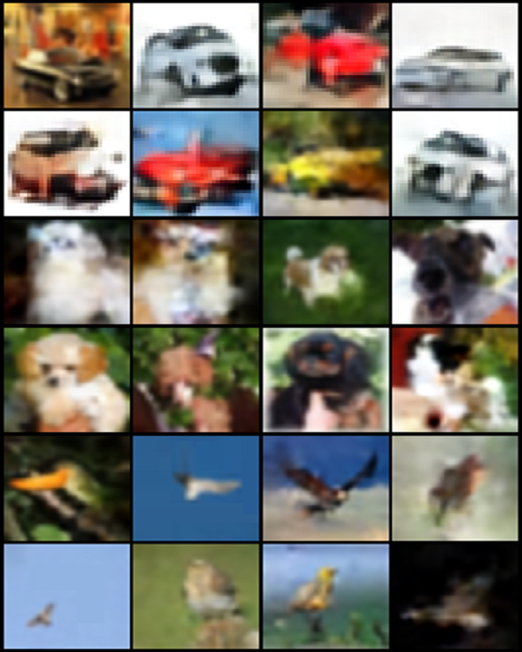}
        \label{fig:cifar10_gen}
    \end{subfigure}
    \hfill
    \begin{subfigure}[b]{0.23\textwidth}  
        \centering 
        \includegraphics[width=\textwidth]{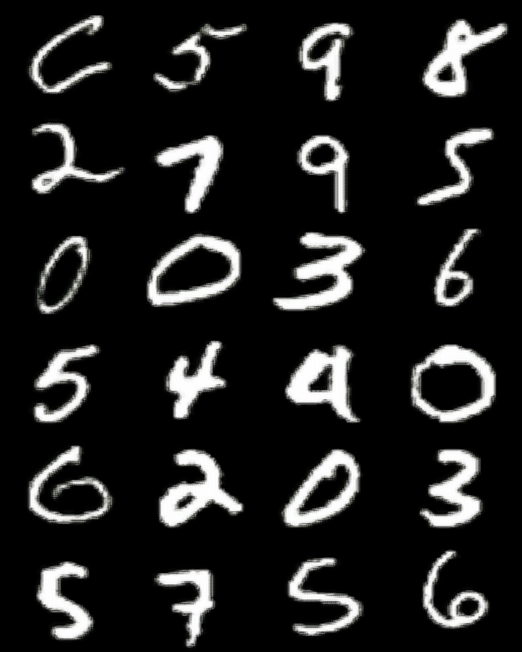}
        \label{fig:mnist_gen}
    \end{subfigure}
    
    \raisebox{5em}{\rotatebox{90}{VQVAE-2}}
    \begin{subfigure}[b]{0.23\textwidth}
        \centering
        \includegraphics[width=\textwidth]{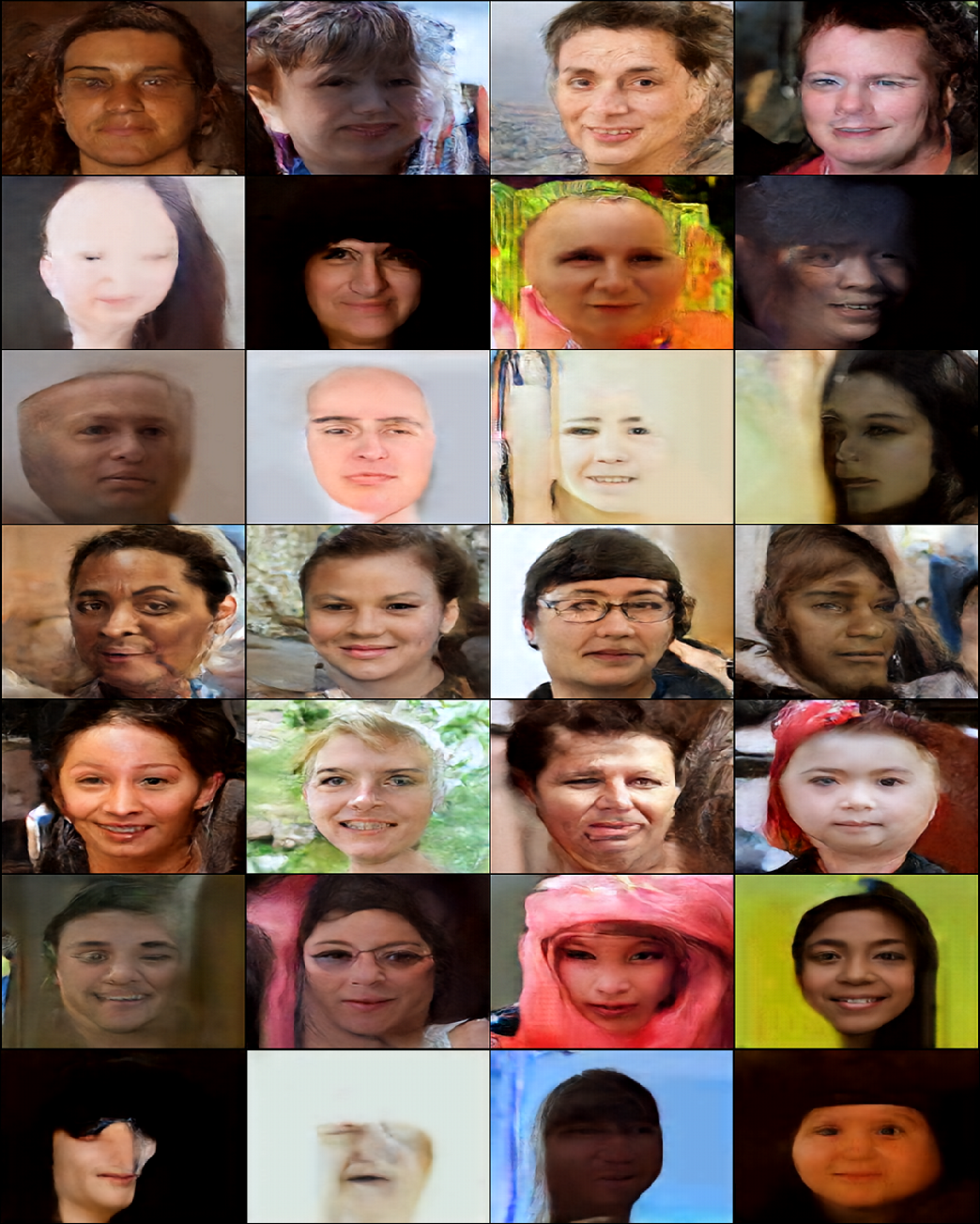}
        \caption{{\small FFHQ}}
        \label{fig:ffhq_vqvae_gen}
    \end{subfigure}
    \hfill
    \begin{subfigure}[b]{0.23\textwidth}  
        \centering 
        \includegraphics[width=\textwidth]{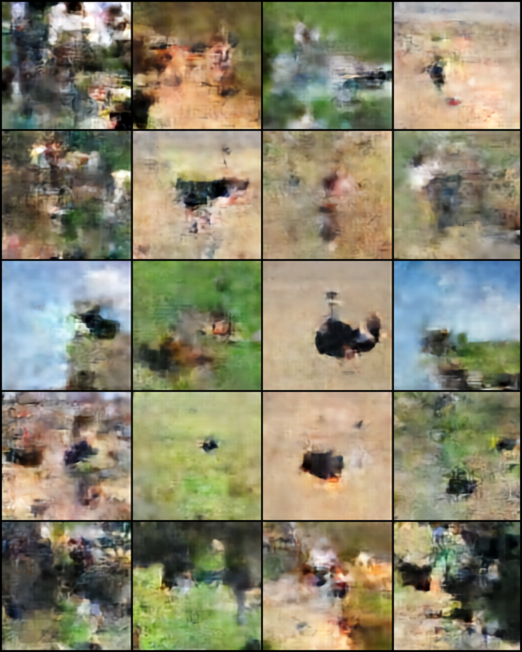}
        \caption{{\small Imagenet}}
        \label{fig:imagenet_vqvae_gen}
    \end{subfigure}
    \hfill
        \begin{subfigure}[b]{0.23\textwidth}
        \centering
        \includegraphics[width=\textwidth]{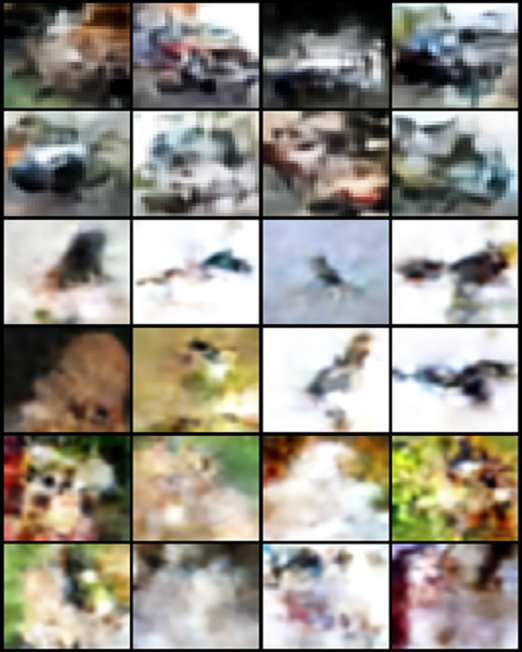}
        \caption{{\small CIFAR10}}
        \label{fig:cifar10_vqvae_gen}
    \end{subfigure}
    \hfill
    \begin{subfigure}[b]{0.23\textwidth}  
        \centering 
        \includegraphics[width=\textwidth]{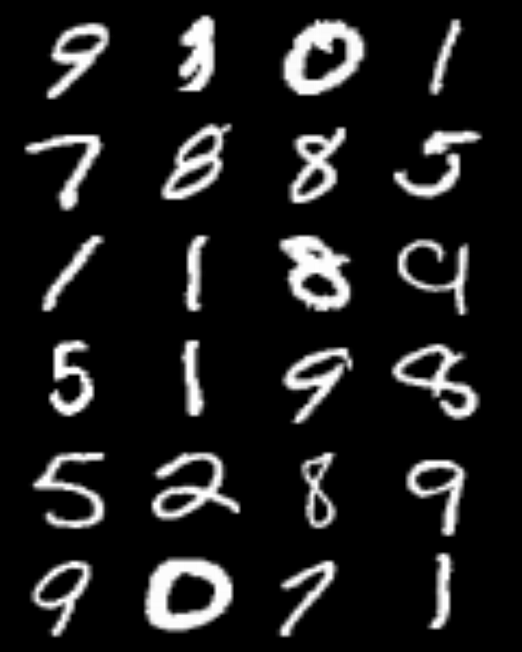}
        \caption{{\small MNIST}}
        \label{fig:mnist_vqvae_gen}
    \end{subfigure}
    \caption{Random samples generated by HR-VQVAE and VQVAE-2.}
    \label{fig:multi_generation}
\end{figure}

Fig.~\ref{fig:multi_generation} presents random samples generated by HR-VQVAE and VQVAE-2. It can be seen that the proposed HR-VQVAE can generate more realistic samples showing the superiority of our model.
Table~\ref{tbl:generation_error} reports the FID results for generated samples with different models. HR-VQVAE reaches lower FID than the baseline models (VQVAE and VQVAE-2). Furthermore, on FFHQ HR-VQVAE (with PixelCNN for sampling), shows a better performance (17.45) than VDVAE~\cite{child2020very} and VQGAN~\cite{esser2021taming} (with PixelCNN for sampling) which reported FIDs 28.50 and 21.93, respectively, but fails against VQGAN (with Transformer~\cite{katharopoulos2020transformers} for sampling) with FID 11.44 which uses a pre-trained autoregressive Transformer to predict rasterized image tokens on the FFHQ dataset. It is worth noting that when VQGAN uses PixelCNN to generate samples, its efficiency is considerably reduced, raising directions for future work.% By combining the proposed hierarchical model and Transformer instead of PixelCNN, we can achieve even more significant results.

\begin{table}
    \centering
    \begin{tabular}{lcccc}
    \hline\hline
    \multirow{2}{*}{Model} & \multicolumn{4}{c}{Generation evaluation (FID $\downarrow$)}\\
    & FFHQ & ImageNet & CIFAR10 & MNIST\\
    \hline
    VQVAE~\cite{rolfe2016discrete} & 24.93 & 44.76 & 78.90 & 16.69\\
    VQVAE-2~\cite{razavi2019generating} &19.66 & 39.51 & 74.43 & 11.81\\
    HR-VQVAE &\textbf{17.45} & \textbf{35.29} & \textbf{71.38} & \textbf{11.75}\\
    \hline\hline
    \end{tabular}
    \caption{Generation results using HR-VQVAE, VQVAE-2 and VQVAE.}
    \label{tbl:generation_error}
\end{table}

\section{Conclusion}
\label{sec:conclusion}
In this paper, we proposed a novel multi-layer variational autoencoder method for image modeling that we call HR-VQVAE.
The model learns discrete representations in an iterative and hierarchical fashion.
The loss function that we introduce to train the model is designed to encourage different layers to encode different aspects of an image.
Through experimental evidence, we show how this model can reconstruct images with a higher level of details than state-of-the-art models with similar complexity.
We also show that we can increase the size of the codebooks without incurring the codebook collapse problem that is observed in methods such as VQVAE and VQVAE-2.
We visualize the internal representations in the model in an attempt to explain its superior performance.
Finally, we show that the hierarchical nature of the codebook design allows to dramatically reduce computation time in decoding.

We believe this model has potential interest for the community both for image reconstruction and generation, particularly in high-load tasks.
This is because i) it dramatically compresses the input samples, ii) each layer captures different levels of abstractions, which allows modeling different aspects of the images in parallel, and iii) the search process is sped up by the hierarchical structure of the codebooks.

%presumably from the context of lower-frequent changes~\cite{barnich2009vibe} in video frames, e.g., background, to detailed frame-to-frame movements (occurring as high-frequent changes~\cite{primus2013segmentation}, e.g., movements).

%\section{Future work}
%\label{sec:future}
%\hl{Such a quantization model can also facilitate frame-by-frame video prediction, as higher layers extract a more general and static concept and lower layers extract the details from images/frames}.

%Although the proposed MLVAE remarkably reduces the dimensionality through hierarchical discrete coding, the reconstructions look barely noisier than the original samples. It would be possible to employ a more perceptual loss function  instead of mean square error over pixels here, but we leave that for future work.

\bibliography{refs}
\newpage
\setcounter{section}{0}
\renewcommand\thesection{\Alph{section}}
\section{Appendix}
\subsection{Architecture Details and Hyperparameters}

%%%%%%%%%%%%%%% VQVAE2 Main %%%%%%%%%%%%%%%%%%%%%%%%%%%%%
\begin{table}
\scriptsize
    \centering
    \begin{tabular}{lcc}
    \hline\hline
    & FFHQ ($1024\times1024$) & Imagenet ($256\times256$) \\
    \hline
    Layers & 3 & 2\\
    Latent layers & $\{128\times128,64\times64,32\times32$\} & \{$64\times64,32\times32$\} \\
    $\beta_i$ & 0.25 & 0.25  \\
    Hidden units & 128 & 128  \\
    Residual units & 64 & 64   \\
    Codebook size & 64 & 64  \\
    Codebook dimension & 64 & 64  \\
    Num. of codewords in layers & \{512,512,512\} & \{512,512,512\} \\
        Num. codewords needed for each pixel & $3\times512$ & $3\times512$  \\
    Encoder Conv filter size & 3 & 3  \\
    Optimizer & Adam & Adam   \\
    Training steps & 304741 & 2207444 \\
    Polyak EMA decay  & 0.9999 & 0.9999  \\
    Learning rate  & $2\times{10}^{-4}$  & $2\times{10}^{-4}$ \\
    \hline\hline
    \end{tabular}
    \caption{Hyper parameters of original VQVAE encoder and decoder used for experiments in the original paper.}
    \label{tbl:hyper_parameters_vqvae_main}
\end{table}
%%%%%%%%%%%%%%%%%%%%%%%%%%%%%%%%%%%%%%%%%%%%%%%%%%%%%%%%
%\\Table~\ref{tbl:input_data} presents a general information about the hyper parameters used in the HR-VQVAE and VQVAE-2 training and testing processes.
\begin{table}
\scriptsize
    \centering
    \begin{tabular}{lcccc}
    \hline\hline
    & FFHQ & Imagenet & CIFAR10 & MNIST\\
    \hline
    Num. of training samples & 60,000 & 1,281,167 & 50,000 & 60,000\\
    Num. of test samples & 10,000 & 100,000 & 10,000 & 10,000 \\
    Input size & $256\times256\times3$ &$128\times128\times3$ & $32\times32\times3$ & $28\times28\times1$  \\
    Batch size & 32 & 64 & 128 & 128 \\
    Number of epochs & 400 & 400 & 999 & 999 \\
    Training steps & 875,000 & 8,007,293  & 390,234 & 468,281 \\
    \hline\hline
    \end{tabular}
    \caption{Input data details for HR-VQVAE and VQVAE2 encoder decoder.}
    \label{tbl:input_data}
\end{table}
%\\The details of hyper-parameters used for training the HR-VQVAE and VQVAE2 encoder and decoder networks of experiments are reported in Tables~\ref{tbl:hyper_parameters_hrvqvae} and Table~\ref{tbl:hyper_parameters_vqvae2}. 

\begin{table}
\scriptsize
    \centering
    \begin{tabular}{lcccc}
    \hline\hline
    & FFHQ & Imagenet & CIFAR10 & MNIST\\
    \hline
    Layers & 3 & 3 & 3 & 3\\
    Latent layers & $32\times32$ & $32\times32$ & $16\times16$ & $16\times16$ \\
    $\beta_i$ & 0.25 & 0.25 & 0.25 & 0.25 \\
    Hidden units & 128 & 128 & 64 & 64 \\
    Residual units & 64 & 64 & 64 & 64  \\
    Codebook size & 8 & 8 & 8 & 4 \\
    Codebook dimension & 32 & 32 & 16 & 8 \\
    Num. of codewords in layers & \{8,64,512\} & \{8,64,512\} &\{8,64,512\} & \{4,16,64\}\\
        Num. codewords needed for each pixel & $3\times8$ & $3\times8$ & $3\times8$ & $3\times4$   \\
    Encoder Conv filter size & 3 & 3 & 3 & 3  \\
    Optimizer & Adam & Adam & Adam & Adam   \\
    Polyak EMA decay  & 0.9 & 0.9 & 0.9 & 0.9 \\
    Learning rate  & $3\times{10}^{-3}$  & $3\times{10}^{-3}$ & $3\times{10}^{-3}$ & $3\times{10}^{-3}$\\
    \hline\hline
    \end{tabular}
    \caption{Hyper parameters of 3 layers HR-VQVAE encoder and decoder used for experiments.}
    \label{tbl:hyper_parameters_hrvqvae}
\end{table}

\begin{table}
\scriptsize
    \centering
    \begin{tabular}{lcccc}
    \hline\hline
    & FFHQ & Imagenet & CIFAR10 & MNIST\\
    \hline
    Layers & \tiny(Bottom, Mid., Top) & \tiny(Bottom, Mid., Top) & \tiny(Bottom, Mid., Top) & \tiny(Bottom, Mid., Top)\\
    Latent layers & \tiny$\{32\times32, 16\times16, 8\times8\} $ & \tiny$\{32\times32, 16\times16, 8\times8\} $ & \tiny$\{16\times16, 8\times8,  4\times4\} $ & \tiny$\{16\times16, 8\times8,  4\times4\}$  \\
    $\beta_i$ & 0.25 & 0.25 & 0.25 & 0.25 \\
    Hidden units & 128 & 128 & 64 & 64 \\
    Residual units & 64 & 64 & 64 & 64  \\
    Codebook size & 512 & 512 & 512 & 64 \\
    Codebook dimension & 32 & 32 & 16 & 8 \\
    Num. of codewords in layers & \{512,512,512\} & \{512,512,512\} &\{512,512,512\} & \{64,64,64\}\\
    Num. codewords needed for each pixel & $3\times512$ & $3\times512$ & $3\times512$ & $3\times64$   \\
    Encoder Conv filter size & 3 & 3 & 3 & 3  \\
    Optimizer & Adam & Adam & Adam & Adam   \\
    Polyak EMA decay  & 0.9 & 0.9 & 0.9 & 0.9 \\
    Learning rate  & $3\times{10}^{-3}$  & $3\times{10}^{-3}$ & $3\times{10}^{-3}$ & $3\times{10}^{-3}$\\
    \hline\hline
    \end{tabular}
    \caption{Hyper parameters of 3 layers VQVAE2 encoder and decoder used for experiments.}
    \label{tbl:hyper_parameters_vqvae2}
\end{table}

\begin{table}
\scriptsize
    \centering
    \begin{tabular}{lcccc}
    \hline\hline
    & FFHQ & Imagenet & CIFAR10 & MNIST\\
    \hline
    Layers & 3 & 3 & 3 & 3\\
    Latent layers & $32\times32$ & $32\times32$ & $16\times16$ & $16\times16$ \\
    Batch size & 32 & 128 & 512 & 512 \\
    Hidden units & 128 & 128 & 64 & 64 \\
    Residual units & 64 & 64 & 64 & 64  \\
    Attention layers & 4 & 4 & 4 & 4 \\
    Attention head & 8 & 8 & 8 & 8 \\
    Conv. Filter size & 5 & 5 & 5 & 5\\
    Dropout & 0.15 & 0.15 & 0.1 & 0.1  \\
    Output stack layers & 0 & 0 & 0 & 0   \\
    Polyak EMA decay  & 0.99 & 0.99 & 0.99 & 0.99\\
    \hline\hline
    \end{tabular}
    \caption{Hyper parameters of 3 layers HR-VQVAE autoregressive prior networks used for experiments.}
    \label{tbl:pixelCNN_hrvqvae}
\end{table}

\begin{table}
\scriptsize
    \centering
    \begin{tabular}{lcccc}
    \hline\hline
    & FFHQ & Imagenet & CIFAR10 & MNIST\\
    \hline
    Layers  & (Bottom, Mid., Top) & (Bottom, Mid., Top) & (Bottom, Mid., Top) & (Bottom, Mid., Top)\\
    Latent layers & \tiny$\{32\times32, 16\times16, 8\times8\} $ & \tiny$\{32\times32, 16\times16, 8\times8\} $ & \tiny$\{16\times16, 8\times8,  4\times4\} $ & \tiny$\{16\times16, 8\times8,  4\times4\}$  \\
    Batch size & \{32, 64, 128\} & \{64, 128, 256\} & 512 & 512 \\
    Hidden units & 128 & 128 & 64 & 64 \\
    Residual units & 64 & 128 & 64 & 64  \\
    Attention layers & \{0, 1, 4\} &  \{0, 1, 4\} &  \{0, 1, 4\} &  \{0, 1, 4\} \\
    Attention head & \{-,-,8\} & \{-,-,8\} & \{-,-,8\} & \{-,-,8\} \\
    Conv. Filter size & \{5,5,5\} & \{5,5,5\} & \{5,5,5\} & \{5,5,5\}\\
    Dropout  & \{0.25, 0.3, 0.5\} & \{0.25, 0.3, 0.5\} & \{0.25, 0.3, 0.5\} & \{0.25, 0.3, 0.5\}  \\
    Output stack layers & 0 & 0 & 0 & 0   \\
    Polyak EMA decay  & 0.99 & 0.99 & 0.99 & 0.99\\
    \hline\hline
    \end{tabular}
    \caption{Hyper parameters of 3 layers VQVAE2 autoregressive prior networks used for experiments.}
    \label{tbl:pixelCNN_vqvae2}
\end{table}

Table~\ref{tbl:hyper_parameters_vqvae_main} reports the configuration of VQVAE-2 from the original publication.
These models were trained using Google Cloud TPUv3\footnote{\url{https://cloud.google.com/tpu}} on FFHQ $1024\times1024$ and Imagenet $256\times256$. The original paper reported that such a huge number of parameters were trained using 8 TPU cores and 128 GPUs. 
In order to perform our comparisons with VQVAE and VQVAE-2, we had to retrain the models.
Working in an academic institution, we do not have access to these extensive computational resources.
Consequently, the proposed models are much simpler in terms of number of parameters.
To make a fair comparison, we retrained VQVAE and VQVAE-2 using a configuration that is more similar to ours.

Table~\ref{tbl:input_data} reports information on the input data formats we used in the different data sets.
Table~\ref{tbl:hyper_parameters_hrvqvae} reports the hyperparameters for the proposed method (HR-VQVAE) whereas Table~\ref{tbl:hyper_parameters_vqvae2} reports the hyperparameters that we have used for VQVAE-2.
Similarly, Table~\ref{tbl:pixelCNN_hrvqvae} and Table~\ref{tbl:pixelCNN_vqvae2} report the configuration of the autoregressive prior networks used for image generation in HR-VQVAE and VQVAE-2, respectively.

\subsection{HR-VQVAE: hierarchy and codebook access}
\begin{figure}[h!]
    \centering
    \includegraphics[width=\textwidth]{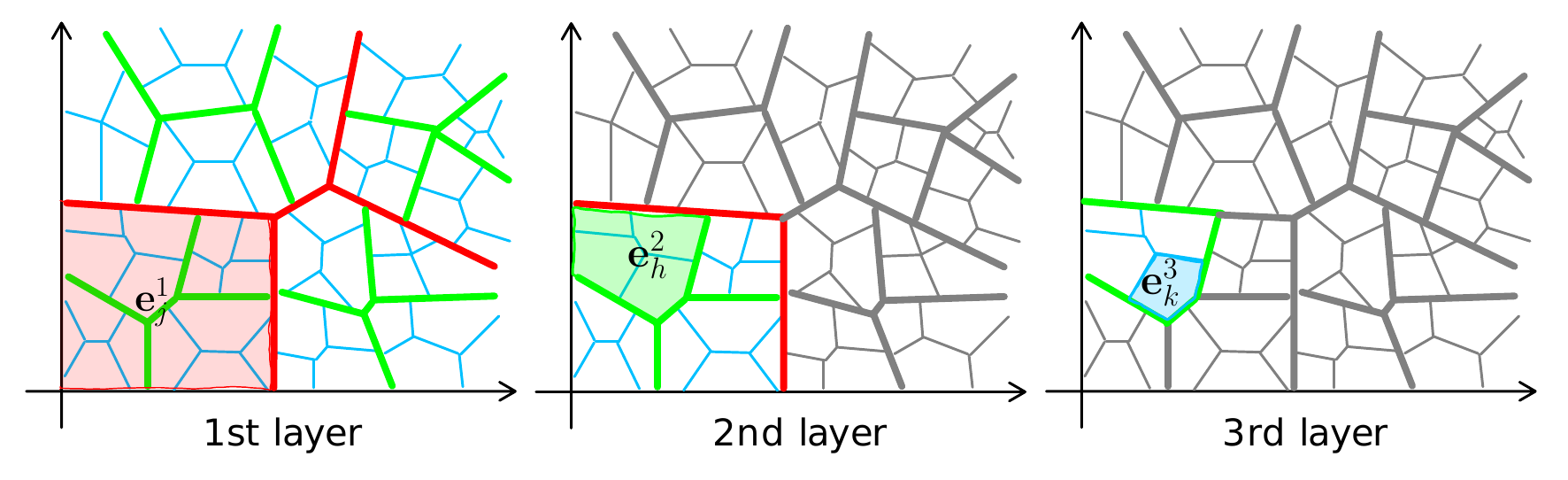}
    \caption{Illustration of the hierarchical codebooks in HR-VQVAE and the speed up in codebook access. }
    \label{fig:hr-vqvae_hierarchical_codebooks}
\end{figure}

Figure~\ref{fig:hr-vqvae_hierarchical_codebooks} illustrates the idea behind the hierarchical codebooks in the proposed method. 
In the example the model has three layers and codebooks of size 4. The plots show layer 1 (left), layer 2 (middle) and layer 3 (right). Also, different color codes are used for codebooks in the layers (red, green and blue, respectively). When a codeword is selected in a certain layer, this limits the codebooks available in the following layers. This is illustrated by graying out the options that are no longer available. In the example, at each layer, the model only has 4 choices, even though the resolution of the last layer has a total of $4^3=64$ codewords. This speeds up inference significantly. Finally, note that each layer only models the residual between the original representation and the representation obtained by combining the codewords at previous layers. This gives the name hierarchical residual learning VQVAE to our model.

\clearpage

\section{Additional samples}
This section reports additional samples generated by the various models we have tested. Refer to the figure caption for more information.
\begin{figure}
    \centering
    \includegraphics[width=0.8\textwidth]{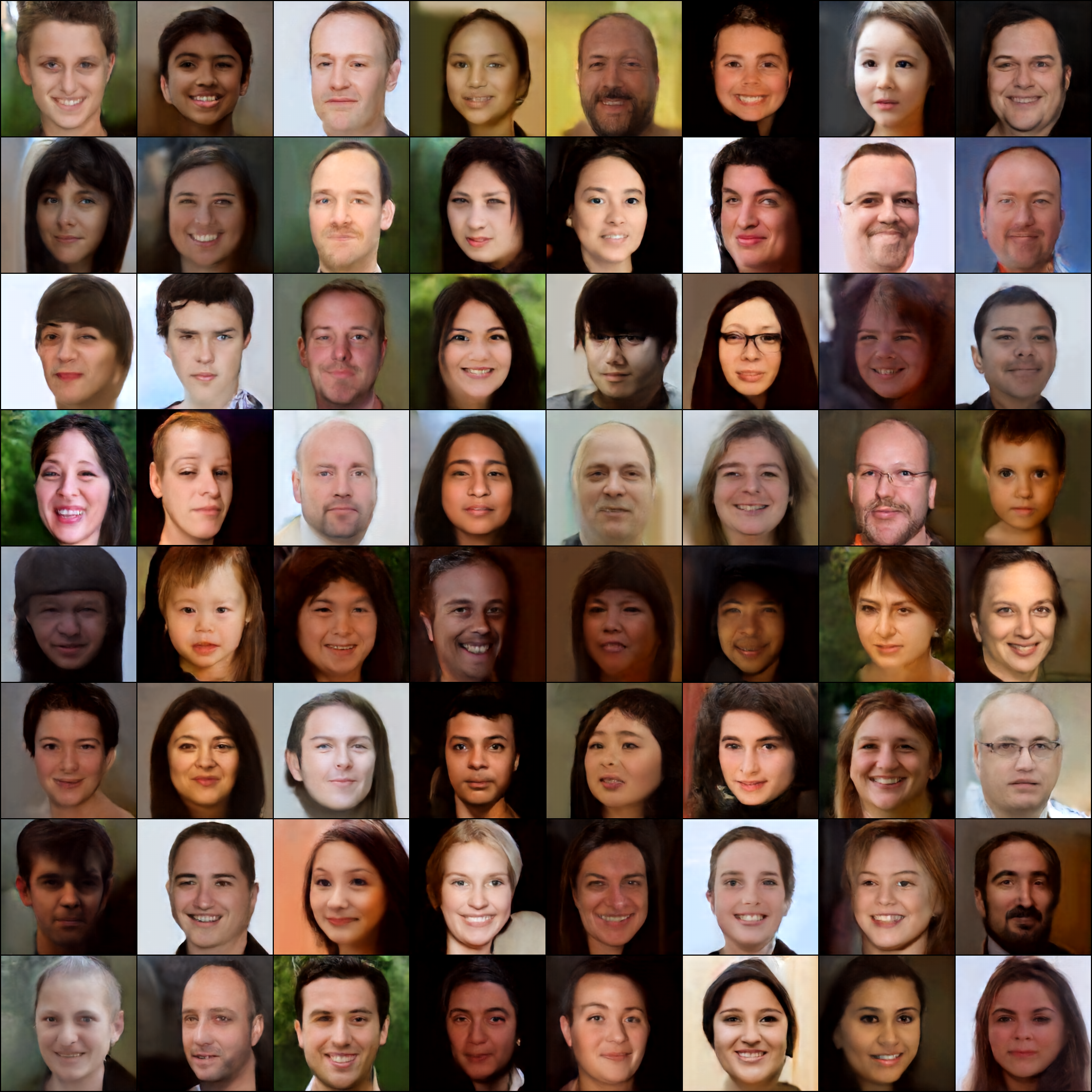}
    \caption{Random samples generated by HR-VQVAE on FFHQ $256\times256$ dataset.}
    \label{fig:hrvqvae_gen_ffhq_supp}
\end{figure}

\begin{figure}
    \centering
    \includegraphics[width=0.8\textwidth]{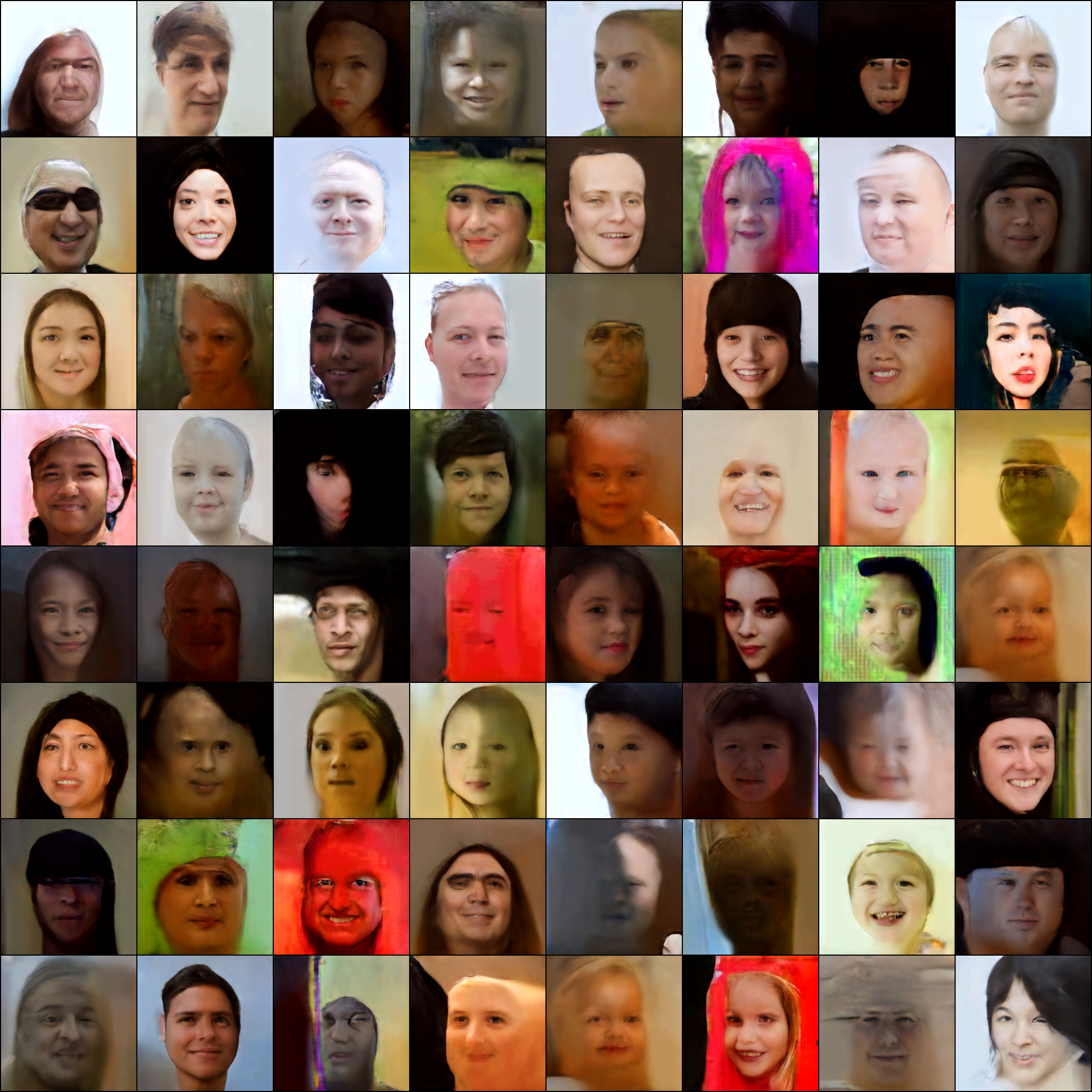}
    \caption{Random samples generated by VQVAE2 on FFHQ $256\times256$ dataset.}
    \label{fig:vqvae2_gen_ffhq_supp}
\end{figure}

\begin{figure}
    \centering
    \includegraphics[width=0.8\textwidth]{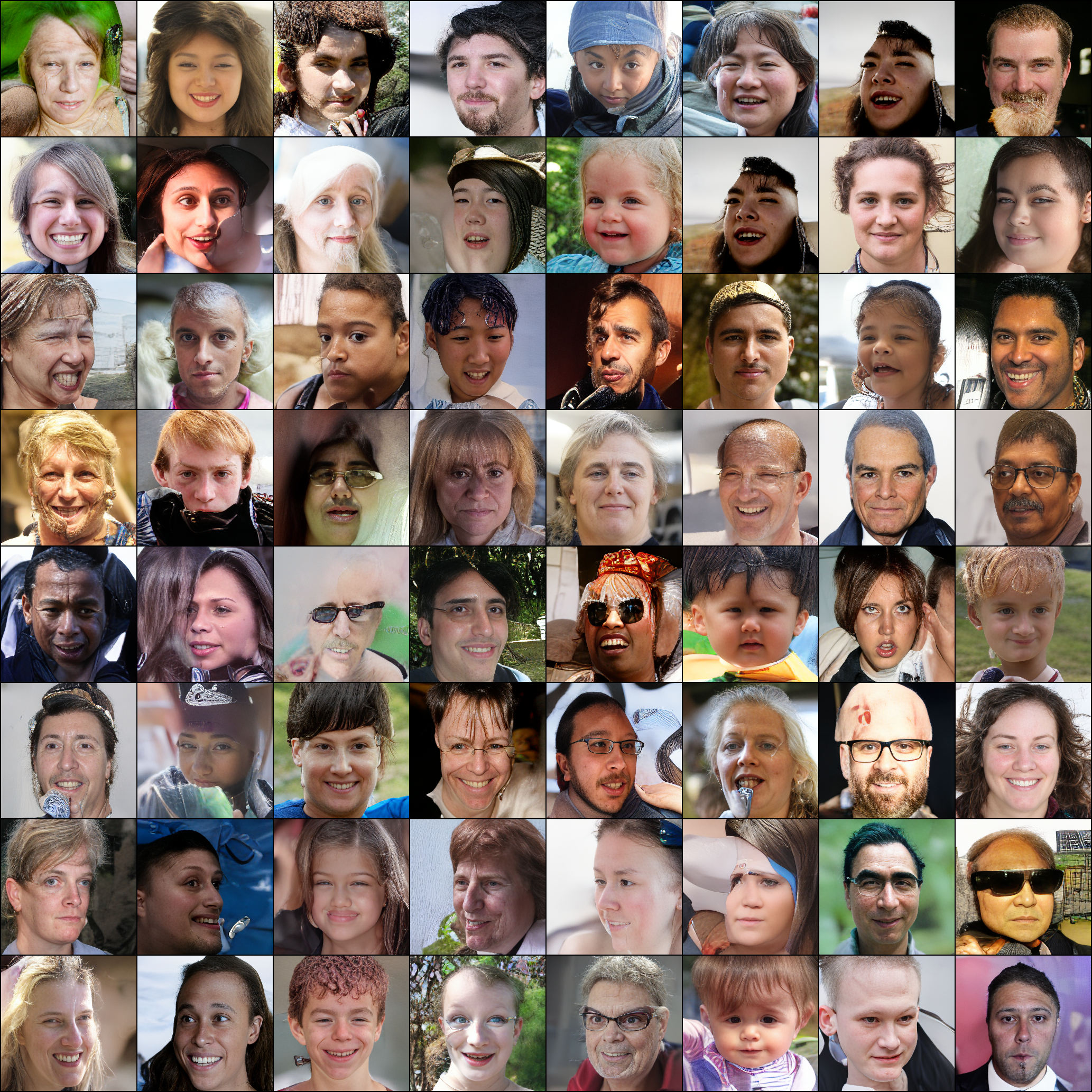}
    \caption{Random samples generated by VQGAN on FFHQ $256\times256$ dataset.}
    \label{fig:vqgan_gen_ffhq_supp}
\end{figure}

\begin{figure}
    \centering
    \includegraphics[width=0.6\textwidth]{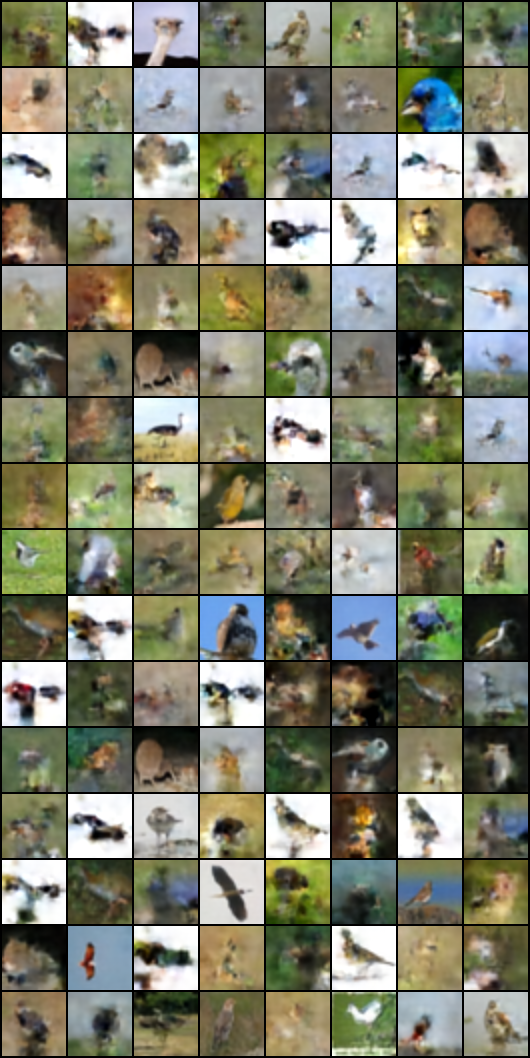}
    \caption{Class conditional random samples generated by HR-VQVAE on CIFAR10-Bird dataset.}
    \label{fig:hrvqvae_gen_cifar10_supp}
\end{figure}

\begin{figure}
    \centering
    \includegraphics[width=0.8\textwidth]{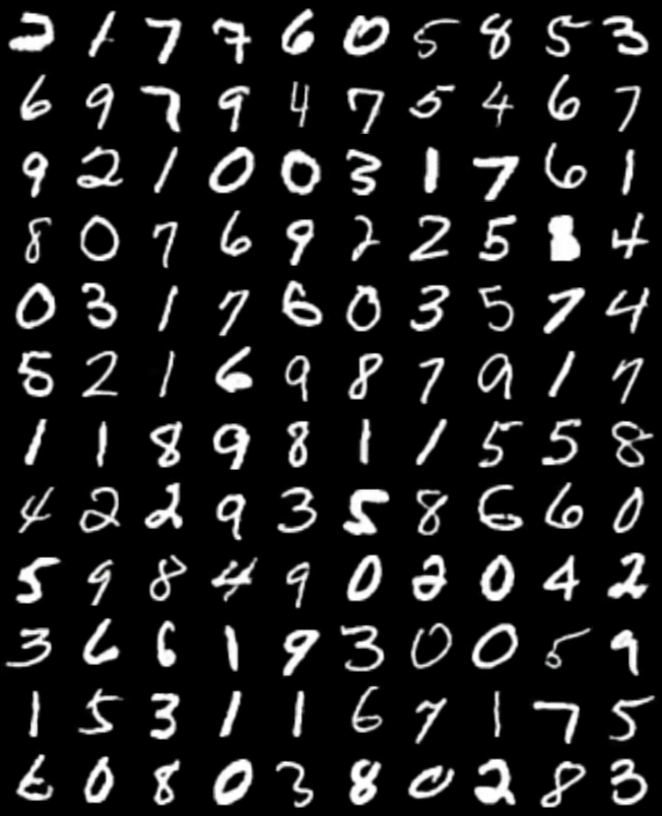}
    \caption{Random samples generated by HR-VQVAE on MNIST dataset.}
    \label{fig:hrvqvae_gen_mnist_supp}
\end{figure}
\end{document}